\documentclass[12pt]{article}
\usepackage{amsmath}
\usepackage{amsthm}
\usepackage{amsfonts}
\usepackage{amssymb}
\usepackage{algorithm}
\usepackage{algpseudocode}
\usepackage{graphicx,psfrag,epsf}
\usepackage{enumerate}
\usepackage{natbib}
\usepackage{url} 
\usepackage[english]{babel}
\usepackage[autostyle]{csquotes}
\usepackage{tabularx}
\usepackage{booktabs}
\usepackage{caption}
\usepackage{mathtools}
\usepackage{bm}
\usepackage{bbm}

\allowdisplaybreaks

\newtheorem{definition}{Definition}
\newtheorem{theorem}{Theorem}
\newtheorem{proposition}{Proposition}
\newtheorem{lemma}{Lemma}
\newtheorem{corollary}{Corollary}
\newtheorem{remark}{Remark}
\newtheorem{example}{Example}
\newtheorem{assumption}{Assumption}

\DeclareMathOperator*{\argmax}{arg\,max}

\newcommand{\blind}{1}

\begin{document}

\def\spacingset#1{\renewcommand{\baselinestretch}%
{#1}\small\normalsize} \spacingset{1}


\if1\blind
{
  \title{\bf Generalized Power Priors for Improved Bayesian Inference with Historical Data}
  \author{Masanari Kimura\thanks{Research Fellow, School of Mathematics and Statistics, The University of Melbourne} \ 
  and
  Howard Bondell\thanks{Professor, School of Mathematics and Statistics, The University of Melbourne}
    }
  \maketitle
} \fi

\if0\blind
{
  \bigskip
  \bigskip
  \bigskip
  \begin{center}
    {\LARGE\bf Generalized Power Priors for Improved Bayesian Inference with Historical Data}
\end{center}
  \medskip
} \fi

\bigskip
\begin{abstract}
The power prior is a class of informative priors designed to incorporate historical data alongside current data in a Bayesian framework.
It includes a power parameter that controls the influence of historical data, providing flexibility and adaptability.
A key property of the power prior is that the resulting posterior minimizes a linear combination of KL divergences between two pseudo-posterior distributions: one ignoring historical data and the other fully incorporating it.
We extend this framework by identifying the posterior distribution as the minimizer of a linear combination of Amari’s $\alpha$-divergence, a generalization of KL divergence.
We show that this generalization can lead to improved performance by allowing for the data to adapt to appropriate choices of the $\alpha$ parameter. Theoretical properties of this generalized power posterior are established, including behavior as a generalized geodesic on the Riemannian manifold of probability distributions, offering novel insights into its geometric interpretation.
\end{abstract}

\noindent%
\vfill

\newpage
\spacingset{1.45} 

\section{Introduction}
\label{sec:introduction}
Scientific investigations in areas such as early-phase
clinical trials, rare-disease studies and high-dimensional omics often suffer from data that are sparse, noisy or otherwise unrepresentative of the target population.
Concurrently, investigators typically possess an expanding archive of historical trials, registries and real-world evidence that can stabilize estimation and sharpen decision making.
The methodological challenge is therefore not whether to borrow information, but how much to borrow and how to calibrate that borrowing so that efficiency gains are not paid for with excessive sensitivity to model–data conflict.
A widely adopted solution is the power prior of \citet{ibrahim2000power}, which raises the historical likelihood to a scalar weight $\xi\in[0,1]$ before combining it with a baseline prior.
Its conceptual clarity and conjugate structure have spurred numerous extensions including joint and normalised forms~\citep{chen2000power,duan2006evaluating}, commensurate and location-adaptive variants~\citep{hobbs2011hierarchical}, effective sample size rules~\citep{ollier2020adaptive}, and scale-transformed approaches for discordant endpoints~\citep{alt2023scale}.
The theoretical underpinning is elegant:
\citet{Ibrahim01032003} showed that the power-prior
posterior minimizes a convex combination of two
Kullback–Leibler (KL) divergences, one anchored at the
no-borrowing and the other at the full-borrowing posterior.

Despite this progress, existing variants inherit the KL foundation of the original power prior.
While KL optimality is attractive for well-specified models, its heavy tail and asymmetric penalty make the resulting posterior sensitive to outliers, prior–data conflict and model misspecification issues that are accentuated when historical studies differ systematically from contemporaneous trials.
This paper closes that gap by lifting the KL criterion to Amari's $\alpha$-divergence~\citep{amari2009alpha} which is a one-parameter family that interpolates smoothly between the forward and reverse KL limits and encompasses a broad spectrum of robust divergences.
The resulting posterior is shown to be the unique minimizer of a weighted sum of $\alpha$-divergences to the no- and full-borrowing pseudo–posteriors, and retaining the optimization strategy of the classical power prior while injecting a tunable robustness parameter $\alpha$.
Moreover, the solution lies on the
$\alpha$-geodesic connecting the two pseudo-posteriors on the statistical manifold, exposing an information-geometric structure not apparent in previous formulations.

\paragraph{Contributions.}
Our main contributions are summarized as follows.
\begin{itemize}
\item  {\bf Divergence-optimal characterization.}
      We generalize existing power prior and prove that the resulting posterior is a minimizer of a weighted sum of $\alpha$–divergences.
\item  {\bf Theoretical guarantees.}
      We establish (i) global prior–data robustness bounds under
      Huber-type contamination, (ii) shape theorems showing how $\alpha$
      controls uni- versus multi-modality, and (iii) consistency and
      higher-order asymptotics whose leading variance term coincides with that
      of standard weighted likelihood but whose $O(n^{-2})$ correction depends
      on $(\alpha,\xi)$.
\item  {\bf Information-geometric insight.}
      We show that the new posterior traces an $\alpha$-geodesic between two distributions on the statistical manifold.
\item  {\bf Worked examples and case study.}
      We provide fully-worked Bayesian survival analysis of two ECOG melanoma trials that demonstrates practical gains in hazard-ratio estimation and predictive concordance.
\end{itemize}

\paragraph{Organization of the paper}
Section~\ref{sec:background} summarizes the conventional power–prior framework and its KL–optimality result.
Section~\ref{sec:generalization} presents the generalized power prior and corresponding posterior construction, proves its divergence‐optimality.
Section~\ref{sec:examples} works out closed-form posteriors for Gaussian, Beta–Bernoulli and Dirichlet–multinomial models that illustrate how the
new parameter $\alpha$ modulates borrowing.
Section~\ref{sec:theory} develops robustness bounds, shape theorems, consistency, higher-order asymptotics and an information-geometric interpretation.
Practical implications are demonstrated in
Section~\ref{sec:application} through a survival analysis of the ECOG E1684 / E1690 melanoma trials.
Section~\ref{sec:conclusion} concludes with a discussion of possible extensions.
Technical proofs and additional derivations are collected in the supplementary material.

\section{Backgroud}
\label{sec:background}
In Bayesian analysis, it is often the case that one can access historical data $D_0$ in addition to the current data $D$.
The power prior, which is useful in such cases, is a class of informative prior distributions first developed by \citet{ibrahim2000power}.
\begin{definition}[Power prior~\citep{ibrahim2000power}]
    \label{def:power_prior}
    Let the data for the current study be denoted by $D$, and denote the corresponding likelihood function by $L(\bm{\theta} \mid D)$, where $\bm{\theta} \in \Theta$ is a vector of parameters in the parameter space $\Theta$.
    Suppose we have historical data $D_0$ from a similar previous study, and let $L(\bm{\theta} \mid D_0)$ be the corresponding likelihood function.
    Then, the power prior is defined as $\pi(\bm{\theta} \mid D_0, \xi) \propto L(\bm{\theta} \mid D_0)^{\xi}\pi_0(\bm{\theta})$, where $0 \leq \xi \leq 1$ is a scalar parameter and $\pi_0$ is the initial prior for $\bm{\theta}$ before the historical data $D_0$ is observed.
\end{definition}
Using the power prior in Definition~\ref{def:power_prior}, the corresponding posterior distribution of $\bm{\theta}$ is given by $\pi(\bm{\theta} \mid D, D_0, \xi) \propto L(\bm{\theta} \mid D)L(\bm{\theta} \mid D_0)^{\xi}\pi_0(\bm{\theta})$.
One can see that $\xi$ weights the historical data relative to the likelihood of the current study, and thus $\xi$ controls the influence of the historical data.
Considering two extreme cases, we can see that
\begin{itemize}
    \item $\xi = 0$: ignore the historical data $D_0$ as $\pi(\bm{\theta} \mid D, D_0, \xi = 0) \propto L(\bm{\theta} \mid D)\pi_0(\bm{\theta})$, and
    \item $\xi = 1$: give equal importance to $D_0$ and $D$, relative to their sample sizes, as $\pi(\bm{\theta} \mid D, D_0, \xi = 1) \propto L(\bm{\theta} \mid D)L(\bm{\theta} \mid D_0)\pi_0(\bm{\theta})$.
\end{itemize}

Since the power prior is a likelihood function raised to a power, it shares all of the properties of likelihood functions, and therefore has several advantages over other priors.
For example, it can be shown that for many classes of models,
\begin{align*}
    \pi(\bm{\theta} \mid D_0, \xi) \approx \mathcal{N}\left(\hat{\bm{\theta}}_0, \xi^{-1}H^{-1}(\hat{\bm{\theta}})\right),
\end{align*}
where $\hat{\bm{\theta}}_0$ is the mode of the power prior and
\begin{align*}
    H^{-1}(\bm{\theta}) = -\frac{\partial^2}{\partial\bm{\theta}\partial\bm{\theta}^\top}\ln\left(\pi(\bm{\theta} \mid D_0, \xi)\right).
\end{align*}

The power prior has been applied in many fields as the useful informative prior, including survival analysis for clinical trials~\citep{chen1999new,chen2002bayesian,ibrahim2012bayesian}, non-inferiority trials for anti-infective products~\citep{gamalo2014bayesian}, anxiety and depression scale in psychology~\citep{matteucci2014bayesian}, benchmark dose estimation in toxicology~\citep{shao2011bayesian}, randomized therapeutic trials~\citep{rietbergen2011incorporation}, pediatric quality of care~\citep{neelon2010bayesian} and evaluating water quality in environmental science~\citep{duan2006evaluating}.

Moreover, many variants of the power prior have been proposed due to its simplicity of formulation.
\citet{chen2000power} developed the hierarchical prior specification by taking $\xi \in [0, 1]$ random, and proposed a variant $\pi^{\mathrm{JPP}}$ called the joint power prior (JPP).
For example, one can use a beta distribution for $\pi_0(\xi)$.
Another modification of the power prior when $\xi$ is random, which was introduced by \citet{duan2006evaluating}, is called the normalized power prior $\pi^{\mathrm{NPP}}$.
The main difference between $\pi^{\mathrm{JPP}}$ and $\pi^{\mathrm{NPP}}$ is that $\pi^{\mathrm{JPP}}$ specifies a joint prior distribution directly for $(\bm{\theta}, \xi)$ while $\pi^{\mathrm{NPP}}$ first specifies a conditional prior distribution for $\bm{\theta}$ given $\xi$ and then specifies a marginal distribution for $\xi$.
A problem with the JPP $\pi^{\mathrm{JPP}}$ is that they do not directly parameterize the commensurability of the historical data and new data.
To solve this problem, \citet{hobbs2011hierarchical} proposed the location commensurate power prior (LCPP).
Similar ideas include the partial discounting power prior~\citep{chen2001bayesian} and partial borrowing power prior~\citep{ibrahim2012bayesian}.
The adaptive power prior (APP)~\citep{ollier2020adaptive} is based on a criterion constructed using the effective sample size (ESS) and the Hellinger distance.
In this criterion, the ESS is used for checking the maximum desired amount of information, and the Hellinger distance is used for tuning the final balance.
Several variants of power prior are developed for settings where historical and current data involve different data types, such as binary and continuous data.
One of these variants is the scale transformed power prior (straPP)~\citep{alt2023scale}.
The derivation for the straPP is based on the assumption that the standardized parameter values are approximately equal for the historical and current data models and it makes sense to consider such a prior when historical and current data have different outcomes, but nonetheless outcomes that measure related characteristics.
In addition, \citet{alt2023scale} developed the normalized version of the straPP.
Moreover, it is useful to develop a generalization of the straPP that provides a degree of robustness when the assumption of equal standardized parameter values does not hold.
Towards this goal, \citet{alt2023scale} developed a generalized scale transformed power prior (Gen-straPP).

One of the most important properties of the power prior is its optimality under the KL divergence.
This property is given by \citet{Ibrahim01032003}.
\begin{theorem}[Optimality of power prior under KL divergence~\citep{Ibrahim01032003}]
    \label{thm:optimality_under_kl_divergence}
    Let $\bm{\theta} \mapsto p_0(\bm{\theta})$ and $\bm{\theta} \mapsto p_1(\bm{\theta})$ be pseudo posteriors defined as follows.
    \begin{align*}
        p_0(\bm{\theta}) &\coloneqq \pi(\bm{\theta} \mid D, D_0, \xi = 0) = L(\bm{\theta} \mid D)\pi_0(\bm{\theta}), \\
        p_1(\bm{\theta}) &\coloneqq \pi(\bm{\theta} \mid D, D_0, \xi = 1) = L(\bm{\theta} \mid D)L(\bm{\theta} \mid D_0)\pi_0(\bm{\theta}).
    \end{align*}
    Here, the posterior $\bm{\theta} \mapsto g(\bm{\theta}) \coloneqq \pi(\bm{\theta} \mid D, D_0, \xi)$ is a minimizer of the following criterion.
    \begin{align*}
        (1 - \xi) D_{\mathrm{KL}}[g \| p_0] + \xi D_{\mathrm{KL}}[g \| p_1],
    \end{align*}
    where $(p, q) \mapsto D_{\mathrm{KL}}[p \| q]$ is the KL divergence defined as
    \begin{align*}
        D_{\mathrm{KL}}[p \| q] \coloneqq \int_\Theta p(\bm{\theta})\ln\frac{p(\bm{\theta})}{q(\bm{\theta})}d\bm{\theta}.
    \end{align*}
\end{theorem}
Another proof using Lagrangians, different from the one given by \citet{Ibrahim01032003} is provided in Section B of the Supplementary Material, for the sake of unifying the discussion that follows.
While the above theorem guarantees optimality under power prior to KL divergence, it is known to have problems such as a lack of robustness to outliers.
In the following, the optimal form of the posterior distribution under generalized divergence is identified and its properties are investigated.

\section{Generalization}
\label{sec:generalization}
Consider the following generalized divergence.
\begin{definition}[Amari's $\alpha$-divergence~\citep{amari2009alpha}]
    \label{def:alpha_divergence}
    For two probability densities $\bm{\theta} \mapsto p(\bm{\theta})$ and $\bm{\theta} \mapsto q(\bm{\theta})$ on the same space $\Theta$, Amari's $\alpha$-divergence $(p, q) \mapsto D_\alpha[p \| q]$ is given as follows.
    \begin{align*}
        D_\alpha[p \| q] \coloneqq \frac{4}{1 - \alpha^2}\left(1 - \int p(\bm{\theta})^{\frac{1 - \alpha}{2}}q(\bm{\theta})^{\frac{1 + \alpha}{2}} d\bm{\theta} \right),
    \end{align*}
    where $\alpha \neq 1$.
\end{definition}
Here, it is known that Amari's $\alpha$-divergence corresponds to KL divergence and its dual, respectively, at $\alpha \to \pm 1$.

For the generalization of Theorem~\ref{thm:optimality_under_kl_divergence}, consider the following minimization problem.
\begin{align*}
        & \min_{g(\bm{\theta})}\left\{(1 - \xi) D_\alpha[g \| p_0] + \xi D_\alpha[g \| p_1] \right\}, \\
    & \text{subject to} \int_\Theta g(\bm{\theta}) d\bm{\theta} = 1,\ g(\bm{\theta}) \geq 0.
\end{align*}
Let $S_i(g) = \int_\Theta g(\bm{\theta})^{\frac{1 - \alpha}{2}}p_i(\bm{\theta})^{\frac{1 + \alpha}{2}}$ for $i \in \{0, 1\}$.
Then, 
\begin{align*}
    (1 - \xi) D_\alpha[g \| p_0] + \xi D_\alpha[g \| p_1] &= \frac{4}{1 - \alpha^2}\left\{(1 - \xi)\left(1 - S_0(g)\right) + \xi\left(1 - S_1(g) \right)\right\} \\
    &= \frac{4}{1 - \alpha^2}\left\{1 - \left((1 - \xi)S_0(g) + \xi S_1(g)\right)\right\}.
\end{align*}
Minimizing the above is equivalent to maximizing the following.
\begin{align*}
    (1 - \xi) S_0(g) + \xi S_1(g) &= (1 - \xi) \int_\Theta g(\bm{\theta})^{\frac{1 - \alpha}{2}}p_0(\bm{\theta})^{\frac{1 + \alpha}{2}}d\bm{\theta} + \xi \int_\Theta g(\bm{\theta})^{\frac{1 - \alpha}{2}}p_1(\bm{\theta})^{\frac{1 + \alpha}{2}}d\bm{\theta} \\
    &= \int_\Theta g(\bm{\theta})^{\frac{1 - \alpha}{2}}\left\{(1 - \xi)p_0(\bm{\theta})^{\frac{1 + \alpha}{2}} + \xi p_1(\bm{\theta})^{\frac{1 + \alpha}{2}}\right\}d\bm{\theta}.
\end{align*}
Let $A(\bm{\theta}) = (1 - \xi)p_0(\bm{\theta})^{\frac{1 + \alpha}{2}} + \xi p_1(\bm{\theta})^{\frac{1 + \alpha}{2}}$.
Now, the objective is
\begin{align*}
    g^* = \argmax_{g \colon \int_\Theta g(\bm{\theta})d\bm{\theta} = 1} \int_\Theta g(\bm{\theta})^{\frac{1 - \alpha}{2}} A(\bm{\theta}) d\bm{\theta}.
\end{align*}
Introduce a Lagrange multiplier $\lambda$ for the constraint $\int_\Theta g(\bm{\theta}) d\bm{\theta} = 1$, and define the Lagrangian functional
\begin{align*}
    \mathcal{L}(g, \lambda) \coloneqq \int_\Theta g(\bm{\theta})^{\frac{1 - \alpha}{2}}A(\bm{\theta})d\bm{\theta} - \lambda \left\{\int_\Theta g(\bm{\theta})d\bm{\theta} - 1\right\}.
\end{align*}
Then, consider the Gateaux derivative of $\mathcal{L}$ with respect to $g(\bm{\theta})$ and require it to vanish at the optimum as $\frac{\delta \mathcal{L}(g, \lambda)}{\delta g(\bm{\theta})} = 0$.
Since
\begin{align*}
    \frac{\delta}{\delta g(\bm{\theta})}\left(g(\bm{\theta})^{\frac{1 - \alpha}{2}}A(\bm{\theta})\right) &= \frac{1 - \alpha}{2}g(\bm{\theta})^{\frac{1 - \alpha}{2} - 1}A(\bm{\theta}) = \frac{1 - \alpha}{2}g(\bm{\theta})^{-\frac{1 + \alpha}{2}}A(\bm{\theta}),
\end{align*}
the stationarity condition gives
\begin{align*}
    \frac{1 - \alpha}{2}g(\bm{\theta})^{-\frac{1 + \alpha}{2}}A(\bm{\theta}) - \lambda = 0, \quad \Rightarrow \quad  g(\bm{\theta}) = \left\{ \frac{1 - \alpha}{2\lambda}A(\bm{\theta}) \right\}^{\frac{2}{1 + \alpha}}.
\end{align*}
Here, for enforcing normalization, $1 = \int_\Theta g(\bm{\theta}) d\bm{\theta} = \int_\Theta C A(\bm{\theta}) d\bm{\theta}$, where $C = ((1 - \alpha) / 2\lambda)^{2 / (1 + \alpha)}$, and $C = 1 / \int_\Theta A(\bm{\theta})^{\frac{2}{1 + \alpha}} d\bm{\theta}$.
Therefore, the unique minimizer is
\begin{align}
    g^*(\bm{\theta} \mid D) = \frac{A(\bm{\theta})^{\frac{2}{1 + \alpha}}}{\int_\Theta A(\bm{\theta}')^{\frac{2}{1 + \alpha}} d\bm{\theta}'} = \frac{\left\{(1 - \xi)p_0(\bm{\theta})^{\frac{1 + \alpha}{2}} + \xi p_1(\bm{\theta})^{\frac{1 + \alpha}{2}}\right\}^{\frac{2}{1 + \alpha}}}{\int_\Theta \left\{(1 - \xi) p_0(\bm{\theta}')^{\frac{1 + \alpha}{2}} + \xi p_1(\bm{\theta}')^{\frac{1 + \alpha}{2}}\right\}^{\frac{2}{1 + \alpha}} d\bm{\theta}'}. \label{eq:power_sum_posterior}
\end{align}
Denote $g^*(\theta)$ in Eq.~\eqref{eq:power_sum_posterior} as the generalized power posterior.
\begin{definition}[Generalized power posterior]
    The generalized power posterior is defined as Eq.~\eqref{eq:power_sum_posterior}.
\end{definition}
\begin{proposition}[Optimality of generalized power posterior under $\alpha$-divergence]
The generalized power posterior is the minimizer of the linear combination of Amari's $\alpha$-divergence.
\end{proposition}

Next, to identify the prior, consider the decomposition
\begin{align*}
    g^*(\bm{\theta} \mid D) \propto \pi_\alpha(\bm{\theta}) \times L(\bm{\theta} \mid D).
\end{align*}
Then, $\pi_\alpha(\bm{\theta}) = g^*(\bm{\theta}) / L(\bm{\theta} \mid D)$, and
\begin{align}
    g^*(\bm{\theta} \mid D) &\propto \left\{(1 - \xi)p_0(\bm{\theta})^{\frac{1 + \alpha}{2}} + \xi p_1(\bm{\theta})^{\frac{1 + \alpha}{2}}\right\}^{\frac{2}{1 + \alpha}} \nonumber \\
    &= \left\{(1 - \xi) \left(L(\bm{\theta} \mid D)\pi_0(\bm{\theta})\right)^{\frac{1 + \alpha}{2}} + \xi\left(L(\bm{\theta} \mid D)L(\bm{\theta} \mid D_0)\pi_0(\bm{\theta})\right)^{\frac{1 + \alpha}{2}} \right\}^{\frac{2}{1 + \alpha}} \nonumber \\
    &= \left\{L(\bm{\theta} \mid D)^{\frac{1 + \alpha}{2}}\pi_0(\bm{\theta})^{\frac{1 + \alpha}{2}}\left((1 - \xi) + \xi L(\bm{\theta} \mid D_0)^{\frac{1 + \alpha}{2}}\right)\right\}^{\frac{2}{1 + \alpha}} \nonumber \\
    &= L(\bm{\theta} \mid D) \times \pi_0(\bm{\theta}) \times \left\{(1 - \xi) + \xi L(\bm{\theta} \mid D_0)^{\frac{1 + \alpha}{2}}\right\}^{\frac{2}{1 + \alpha}}, \nonumber \\
    \pi_\alpha(\bm{\theta}) &\propto \pi_0(\bm{\theta}) \left\{(1 - \xi) + \xi L(\bm{\theta} \mid D_0)^{\frac{1 + \alpha}{2}}\right\}^{\frac{2}{1 + \alpha}}. \label{eq:power_sum_prior}
\end{align}
Since $\pi_\alpha(\theta)$ does not depend on the current data $D$, it is a valid prior distribution.
Therefore, the following definition can be obtained.
\begin{definition}[Generalized power prior]
    The generalized power prior corresponding to the generalized power posterior is defined as Eq.~\eqref{eq:power_sum_prior}.
\end{definition}

\section{Examples}
\label{sec:examples}

In this section, several explicit examples of generalized power posterior are provided (also see Figure~\ref{fig:explicit_examples} for randomly generated $D$ and $D_0$).
See Section C in the Supplementary Material for detailed derivation.

\begin{example}[Univeriate Gaussian]
    Let $D = \{X_1,\dots,X_n\}$, $X_i \mid \theta \sim \mathcal{N}(\theta, \sigma^2)$ and $D_0 = \{Y_1,\dots,Y_n\}$, $Y_j \mid \theta \sim \mathcal{N}(\theta, \sigma^2)$ be the current and historical data with known $\sigma^2$.
    Also $\pi_0(\theta) = \mathcal{N}(\mu_0, \tau_0^2)$ be the baseline prior.
    Then, for $\alpha \neq \pm 1$ and $\xi \in [0, 1]$, the generalized power posterior $\theta \mapsto g(\theta)$ is given as
    \begin{align*}
       g(\theta) \propto \exp\left\{-\frac{1}{2}\left(\frac{n}{\sigma^2} + \frac{1}{\tau_0^2}\right)\theta + \left(\frac{S_X}{\sigma^2} + \frac{\mu_0}{\tau_0^2}\right)\theta\right\}\left\{1 + C\exp\left(\frac{z S_Y}{\sigma^2}\theta - \frac{z n_0}{2\sigma^2}\theta^2\right)\right\}^{1/z}.
   \end{align*}
    where $z = (1 + \alpha) / 2$, $S_X = \sum^n_{i=1}X_i$, $S_Y = \sum^{n_0}_{j=1}Y_j$, and
    \begin{align*}
        C &= \frac{Q_1}{Q_0}\exp\left[-\frac{z}{2\sigma^2}\sum^{n_0}_{j=1}Y_j^2\right], \\
        Q_0 &= (1 - \xi)\frac{1}{(2\pi\sigma^2)^{\frac{nz}{2}}}\frac{1}{(2\pi\tau_0^2)^{\frac{z}{2}}}, \\
        Q_1 &= \xi\frac{1}{(2\pi\sigma^2)^{\frac{(n + n_0)z}{2}}}\frac{1}{(2\pi\tau_0^2)^{\frac{z}{2}}}.
    \end{align*}
    The normalization constant is determined by integrating with respect to $\theta$.
\end{example}

\begin{figure}[H]
    \centering
    \includegraphics[width=\linewidth]{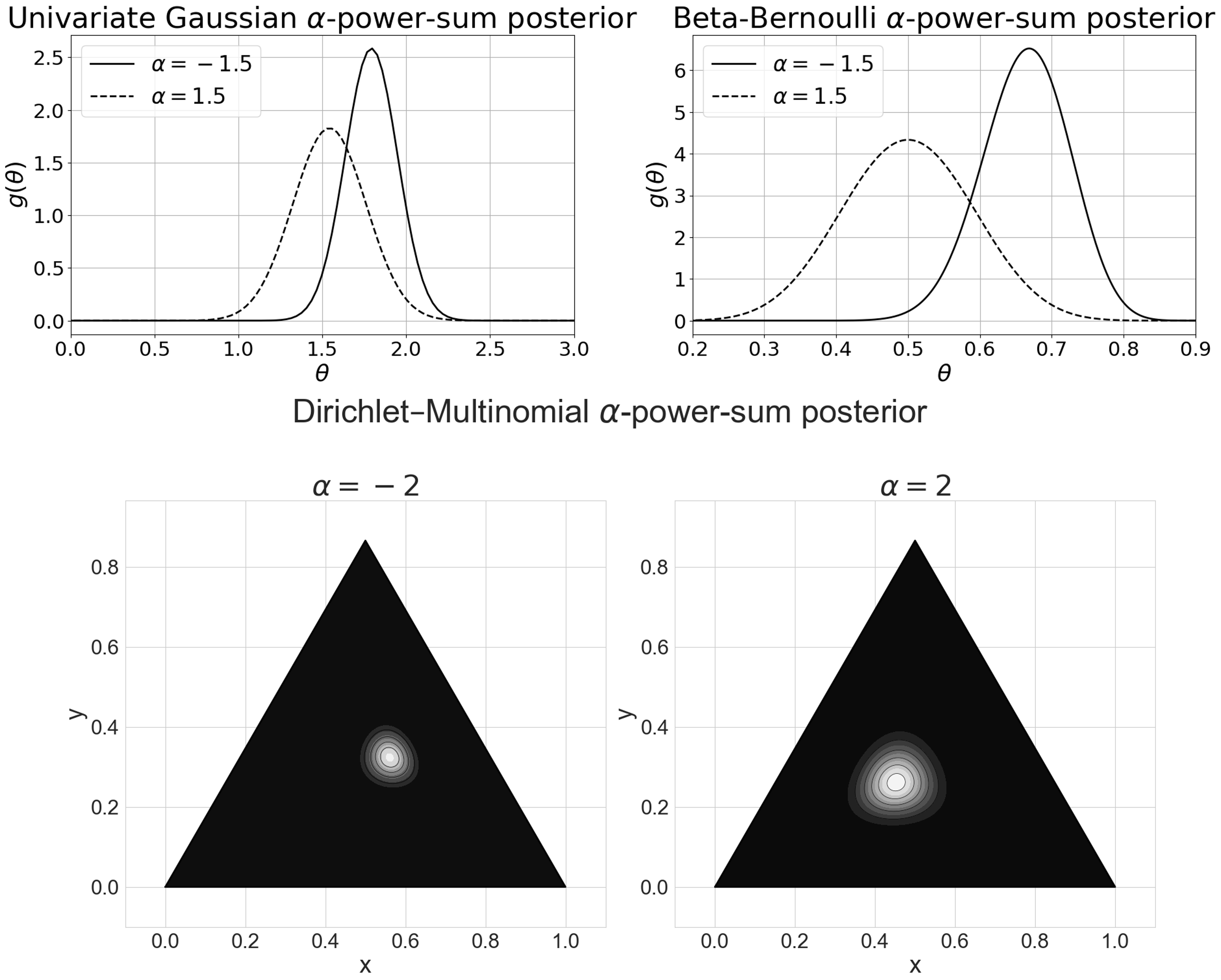}
    \caption{Explicit examples of generalized power posterior. Univariate Gaussian case: $\bar{X} = 1$, $\bar{Y} = 2$, $\sigma^2=1$, $\mu_0=1$ and $\tau_0^2=1$. Beta-Bernoulli case: $\bar{X} = 0.6$, $\bar{Y}=0.7$, $\alpha_0 = 2$ and $\beta_0 = 8$. Dirichlet–Multinomial case: $\bm{\theta} = (20, 15, 15)^\top$ and $\bm{\alpha}_0 = (2.0, 2.0, 2.0)^\top$.}
    \label{fig:explicit_examples}
\end{figure}

\begin{example}[Beta–Bernoulli]
    Let $D = \{X_1,\dots,X_n\}$ with $X_i \in \{0, 1\}$ and $D_0 = \{Y_1,\dots,Y_{n_0}\}$ with $Y_j \in \{0, 1\}$ be the current and historical data from Bernoulli distributions with parameter $\theta \in (0, 1)$.
    Also, suppose that the baseline prior is $\pi_0(\theta) = \mathrm{Beta}(\alpha_0, \beta_0)$ for $\alpha_0, \beta_0 > 0$.
    Then, the corresponding generalized power posterior is given as
    \begin{align*}
        g(\theta) \propto \left\{F(\theta)\right\}^{\frac{1}{z}},
    \end{align*}
    where $z = \frac{1 + \alpha}{2}$, and
    \begin{align*}
        F(\theta) &= \theta^{z(S_X + \alpha_0 - 1)}(1 - \theta)^{z(n - S_X + \beta_0 - 1)} \times \left\{(1 - \xi) + \xi \theta^{z S_Y}(1 - \theta)^{z(n_0 - S_Y)}\right\},
    \end{align*}
    with $S_X = \sum^n_{i=1}X_i$ and $S_Y = \sum^{n_0}_{j=1}Y_j$.
    This is generally not Beta anymore unless $\alpha = 1$ or $\xi=0,1$.
\end{example}

\begin{example}[Dirichlet–Multinomial]
    Let $D = \{X_1,\dots,X_n\}$ with $\sum^n_{i=1 }X_i = n$ and $D_0 = \{Y_1,\dots,Y_{n_0}\}$ with $\sum^{n_0}_{j=1}Y_j = n_0$ from the multinomial distribution having the parameter $\bm{\theta} = (\theta_1,\theta_2,\dots,\theta_k)$.
    Also, suppose that the baseline prior is $\pi_0(\theta) = \mathrm{Dirichlet}(\alpha_{0,1},\alpha_{0,2},\dots,\alpha_{0,k})$.
    Then, the corresponding generalized power posterior is given as
    \begin{align*}
        g(\bm{\theta}) \propto \left\{(1 - \xi)\prod^k_{i=1}\theta_i^{z(X_i + \alpha_{0,i} - 1)} + \xi \prod^k_{j=1}\theta_i^{z(X_i + Y_j + \alpha_{0,j} - 1)} \right\}^{\frac{1}{z}},
    \end{align*}
    where $z = \frac{1 + \alpha}{2}$.
    This does not yield a simple Dirichlet form, unless $\alpha = 1$, that is a non-Dirichlet shape on the simplex.
\end{example}

\begin{remark}[Non-Conjugacy]
    Even if $p_0$ and $p_1$ arise from a conjugate family, the resulting generalized power posterior is typically not in the same conjugate family unless $\alpha = 1$ or $\xi \in \{0, 1\}$.
\end{remark}

\section{Theory}
\label{sec:theory}
In this section, we provide theoretical analysis to verify the behavior of the generalized power posterior.
In Section A in the Supplementary Material, we also provide some auxiliary results.
In addition, see Section B in the Supplementary Material for all necessary proofs in this section.

\subsection{Global Prior Sensitivity and Robustness against Outliers}
\label{sec:theory:global_prior_sensitivity}
One of the important aspects of posterior distributions in Bayesian analysis is its prior sensitivity~\citep{lopes2011confronting,liu2008bayes,muller2012measuring}.
Consider the following contamination model~\citep{huber1992robust,law1986robust,mu2023huber,chen2018robust}.
\begin{definition}[Huber's location shift contamination model~\citep{huber1992robust}]
    \label{def:hubers_contamination_model}
    For the data $D = \{X_1,\dots,X_n\}^n_{i=1}$, suppose that each $X_i$ is independently drawn from the contaminated distribution,
    \begin{align*}
        X_i \sim P_H \coloneqq (1 - \epsilon_H) \mathcal{N}(\theta_0, \sigma^2) + \epsilon_H \mathcal{N}(\theta_H, \sigma^2), 
    \end{align*}
    where $\theta_0$ is the true parameter, $\theta_H$ ($\theta_0 \neq \theta_H$) is the outlier parameter, $\sigma^2 > 0$ is the known variance and $\epsilon \in [0, 1/2]$ is the contamination proportion.
\end{definition}
Under Huber's contamination model, the following robustness analysis is obtained.
\begin{figure}[t]
    \centering
    \includegraphics[width=\linewidth]{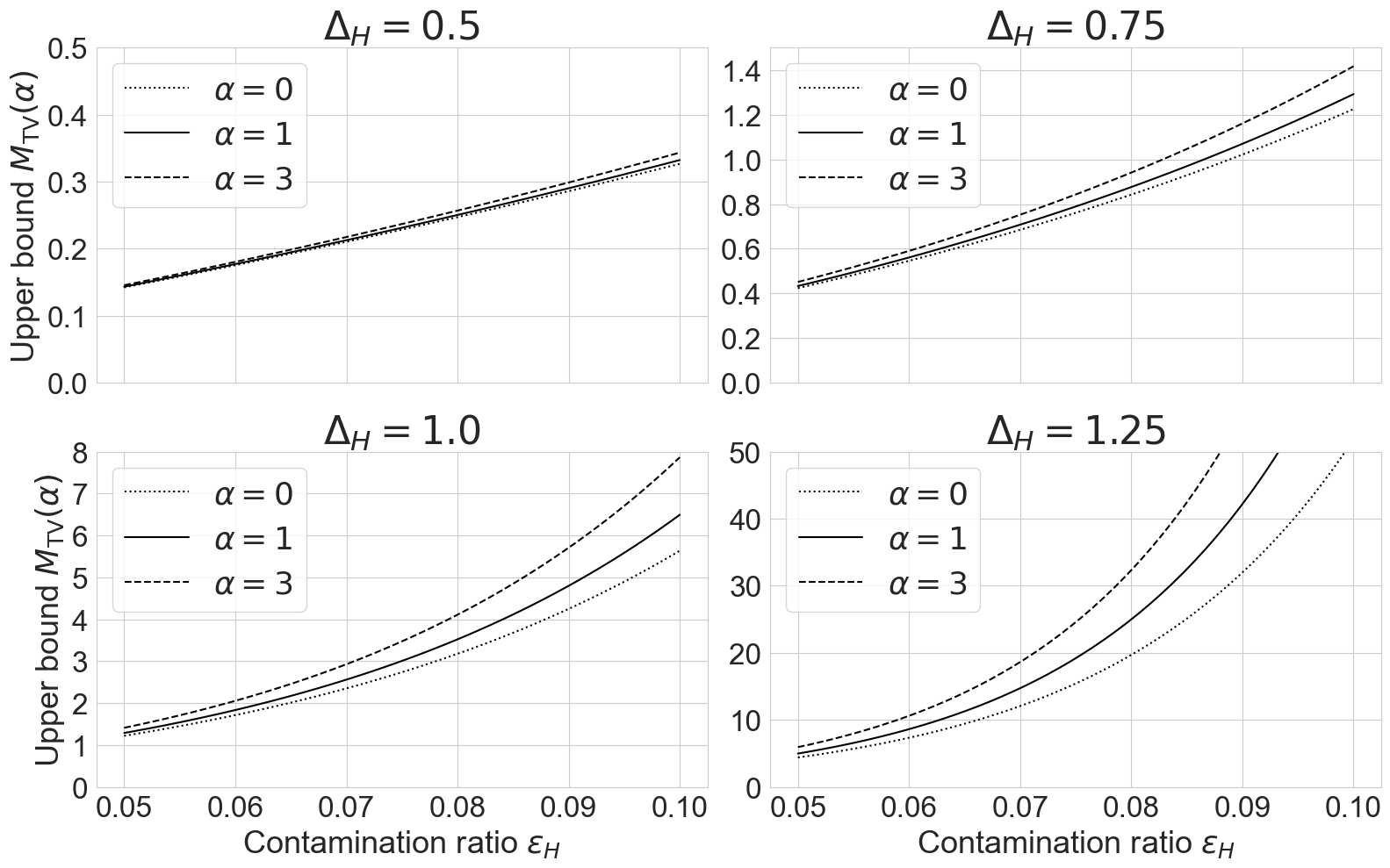}
    \caption{Upper bound $M_{\mathrm{TV}}(\alpha)$ of $d_{\mathrm{TV}}(g^*(\cdot; P_H), g^*(\cdot; P))$ with respect to $\alpha$.}
    \label{fig:tv_bound}
\end{figure}
\begin{figure}[t]
    \centering
    \includegraphics[width=\linewidth]{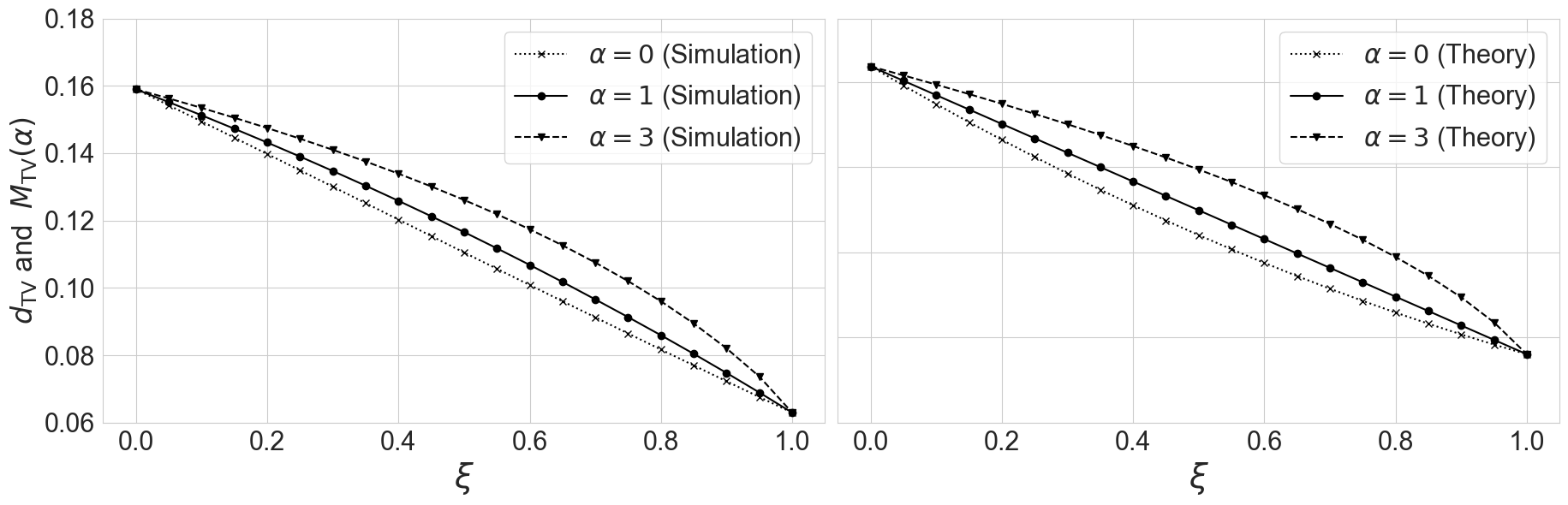}
    \caption{Comparison of empirical $d_{\mathrm{TV}}$ and theoretical bounds.}
    \label{fig:tv_bound_simulation_theory}
\end{figure}
\begin{figure}[t]
    \centering
    \includegraphics[width=0.5\linewidth]{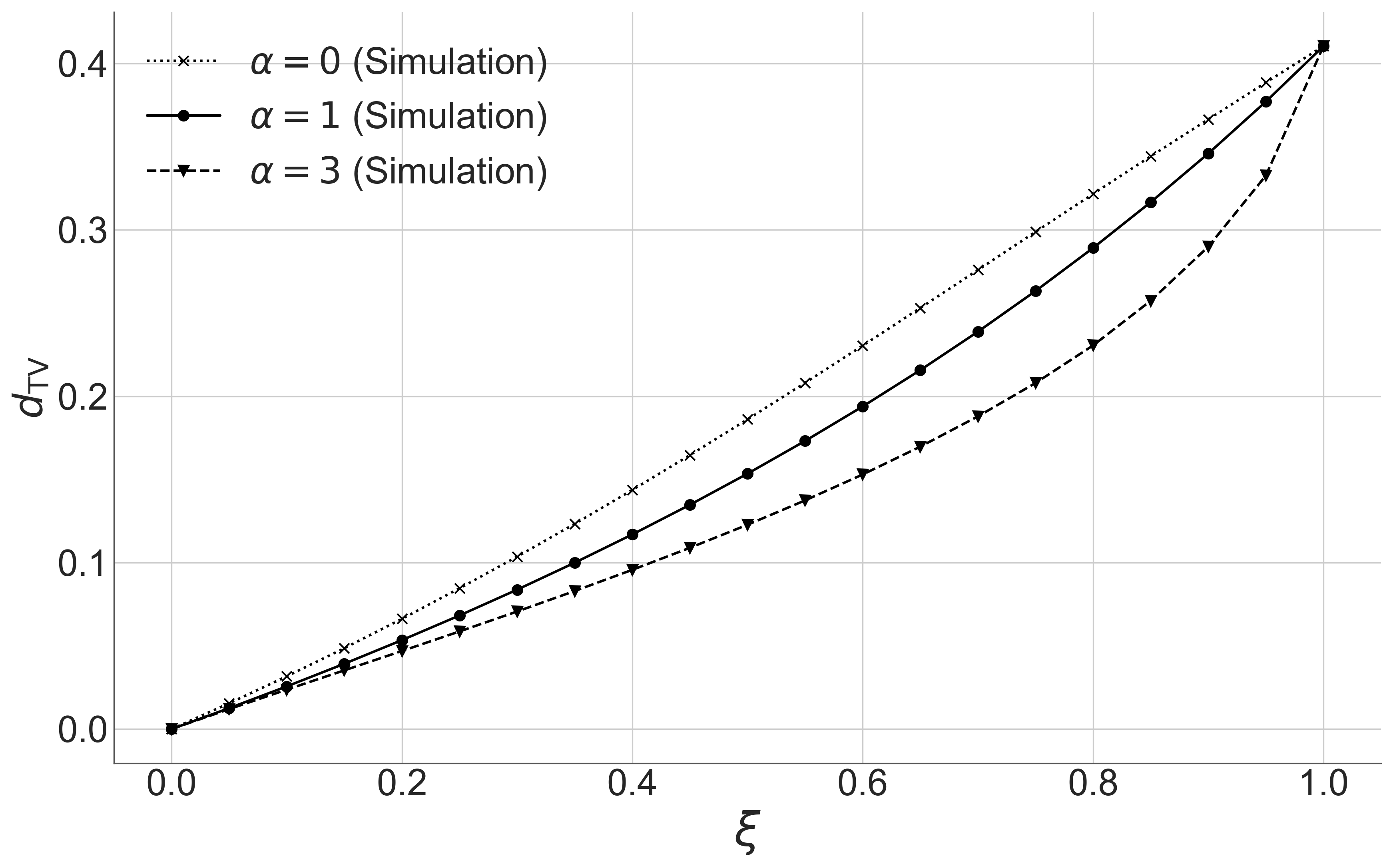}
    \caption{Empirical $d_{TV}$ under reverse contamination.}
    \label{fig:tv_bound_reverse}
\end{figure}
\begin{theorem}[Robustness of generalized power posterior]
    \label{thm:robustness_of_alpha_power_sum_posterior}
    Suppose that the current data $D \sim P_H$ and the historical data $D_0 \sim P$, where $P_H = (1 - \epsilon_H) \mathcal{N}(\theta_0, \sigma^2) + \epsilon_H \mathcal{N}(\theta_H, \sigma^2)$ and $P = \mathcal{N}(\theta_0, \sigma^2)$ are the contaminated and clean distributions.
    Let $p_0$ and $p_1$ be reference densities as
    \begin{align*}
        p_0(\theta) &= L(\theta)\pi_0(\theta),\quad p_1(\theta) = L(\theta)L_0(\theta)\pi_0(\theta), \\
        p_0^F(\theta) &= L^F(\theta)\pi_0(\theta),\quad p_1^F(\theta) = L^F(\theta)L_0(\theta)\pi_0(\theta).
    \end{align*}
    where
    \begin{align*}
        L(\theta) &= \prod^n_{i=1}(1 - \epsilon_H)f_N(X_i ; \theta, \sigma^2) + \epsilon_H f_N(X_i ; \theta_H, \sigma^2), \\
        L_0(\theta) &= \prod^{n_0}_{j=1}f_N(Y_j ; \theta, \sigma^2), \\
        L^F(\theta) &= \prod^n_{i=1} f_N(X_i; \theta, \sigma^2),
    \end{align*}
    $\pi_0$ is the base prior and $f_N(\cdot ; \theta, \sigma^2)$ denotes the density of $\mathcal{N}(\theta, \sigma^2)$.
    Also, write posterior densities as
    \begin{align*}
        g^*(\theta; P_H) &= \frac{\left[(1 - \xi)p_0(\theta)^{\frac{1+\alpha}{2}} + \xi p_1(\theta)^{\frac{1+\alpha}{2}}\right]^{\frac{2}{1+\alpha}}}{\int^\infty_{-\infty}\left[(1 - \xi)p_0(\theta')^{\frac{1+\alpha}{2}} + \xi p_1(\theta')^{\frac{1+\alpha}{2}}\right]^{\frac{2}{1+\alpha}}d\theta'}, \\
        g^*(\theta; P) &= \frac{\left[(1 - \xi)p^F_0(\theta)^{\frac{1+\alpha}{2}} + \xi p^F_1(\theta)^{\frac{1+\alpha}{2}}\right]^{\frac{2}{1+\alpha}}}{\int^\infty_{-\infty}\left[(1 - \xi)p^F_0(\theta')^{\frac{1+\alpha}{2}} + \xi p^F_1(\theta')^{\frac{1+\alpha}{2}}\right]^{\frac{2}{1+\alpha}}d\theta'}.
    \end{align*}
    The total variation distance between these two posteriors satisfies
    \begin{align*}
        d_{\mathrm{TV}}\Bigl(g^*(\cdot;P_H),\,g^*(\cdot;P)\Bigr) &= \frac{1}{2}\int\Bigl| g^*(\theta;P_H) - g^*(\theta;P) \Bigr| d\theta \nonumber \\
        &\leq \frac{1}{2}\Bigl[\Bigl(\frac{R_{\max}}{R_{\min}}\Bigr)^{1/z} - 1 \Bigr],
    \end{align*}
    where
    \begin{align*}
        R(\theta) &=  \frac{(1-\xi)\,\bigl[p_0(\theta)\bigr]^z + \xi\,\bigl[p_1(\theta)\bigr]^z}{(1-\xi)\,\bigl[p_0^F(\theta)\bigr]^z +  \xi\,\bigl[p_1^F(\theta)\bigr]^z}, \\
        R_{\max} &= \sup_{\theta}\;R(\theta), \quad R_{\min} = \inf_{\theta}\;R(\theta).
    \end{align*}
    In particular, the bound goes to 0 when $\epsilon_H\to0$ or $\Delta_H\to0$, and it exhibits the limiting behavior in $\xi\to0$ or $1$ where the dependence on $\alpha$ disappears at the endpoints, and in $\epsilon_H\to0$ or $\Delta_H\to0$ where contamination vanishes.
\end{theorem}
\begin{corollary}
    The total variation distance between two posteriors in Theorem~\ref{thm:robustness_of_alpha_power_sum_posterior} satisfies
    \begin{align*}
        d_{\mathrm{TV}}\Bigl(g^*(\cdot;P_H),\,g^*(\cdot;P)\Bigr) \leq \frac{1}{2}\max\left\{R_{\max}^{1/z} - 1,\ 1 - R_{\min}^{1/z}\right\}.
    \end{align*}
\end{corollary}
\begin{corollary}
    \label{cor:contamination_alpha}
    Under the contamination setting in Theorem~\ref{thm:robustness_of_alpha_power_sum_posterior} and its reverse setting, let
    \begin{align*}
        B(\alpha)=\frac{1}{2}\Biggl[\Bigl(\frac{R_{\max}(\alpha)}{R_{\min}(\alpha)}\Bigr)^{1/z}-1\Biggr],\quad z=\frac{1+\alpha}{2},
    \end{align*}
    where $R(\theta)$ (and hence $R_{\max}(\alpha),\,R_{\min}(\alpha)$) is defined in terms of the likelihoods and the mixing parameter $\xi$.
    Then the following holds:
    \begin{enumerate}
      \item[(i)]  When the current data is contaminated and the historical data is clean, the total--variation bound is an increasing function of $\alpha$.
      \item[(ii)] When the current data is clean and the historical data is contaminated, the corresponding total--variation bound is a decreasing function of $\alpha$.
    \end{enumerate}
\end{corollary}
Figure~\ref{fig:tv_bound} shows the resulting upper bounds with respect to $\alpha$ and $\Delta_H$, with $n=50$, $\xi=0.5$ and $\sigma=1$.
In addition, Figure~\ref{fig:tv_bound_simulation_theory} shows the comparison between empirical total variation distance and theoretical bounds.
For calculating the empirical bound, we generated $D$ and $D_0$ from $\mathcal{N}(0, 1)$ and $\mathcal{N}(1.25, 1)$,  with $n = n_0 = 50$ and $\epsilon_H = 0.05$.
Each point in the above experiments is the mean of 10 trials with different random seeds.
In addition, Figure~\ref{fig:tv_bound_reverse} shows the empirical total variation distance under reverse contamination such that the current data is always clean and the historical data might be contaminated.

\begin{figure}[t]
    \centering
    \includegraphics[width=0.95\linewidth]{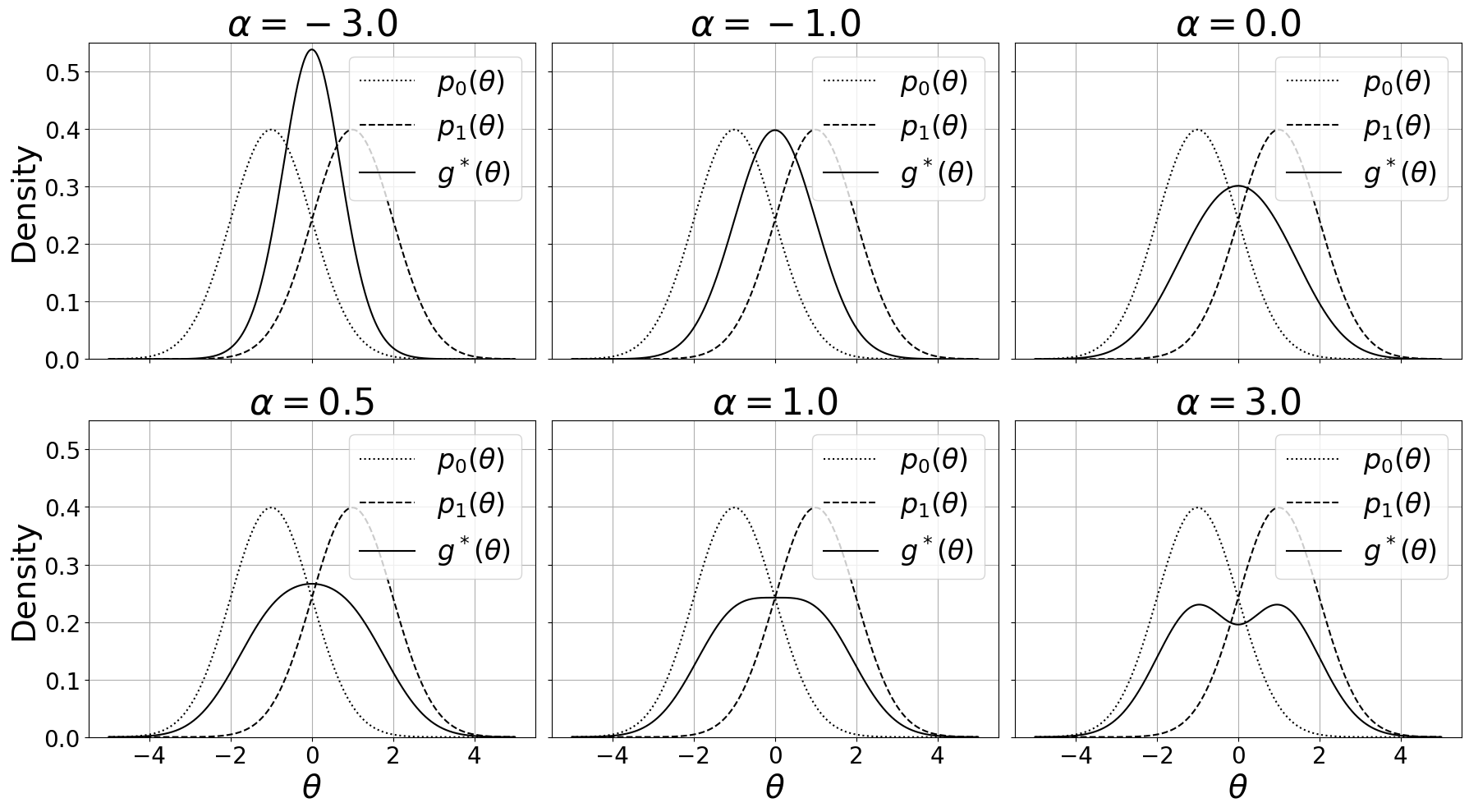}
    \caption{Role of the parameter $\alpha$ for the shape of generalized power posterior (Gaussian distributions with same variance, $\xi = 0.5$). For the case of $\alpha \to \pm 1$, $\alpha = \pm 1 - 10^{-4}$ are used. \label{fig:shape_of_power_sum_posterior}}
\end{figure}

\subsection{Shape of Generalized Power Posterior}
\label{sec:theory:shape_of_power_sum_posterior}
Another interest is the role of parameter $\alpha$ for the shape of generalized power posterior.
To investigate this, consider the following assumptions.

\begin{figure}[H]
    \centering
    \includegraphics[width=0.95\linewidth]{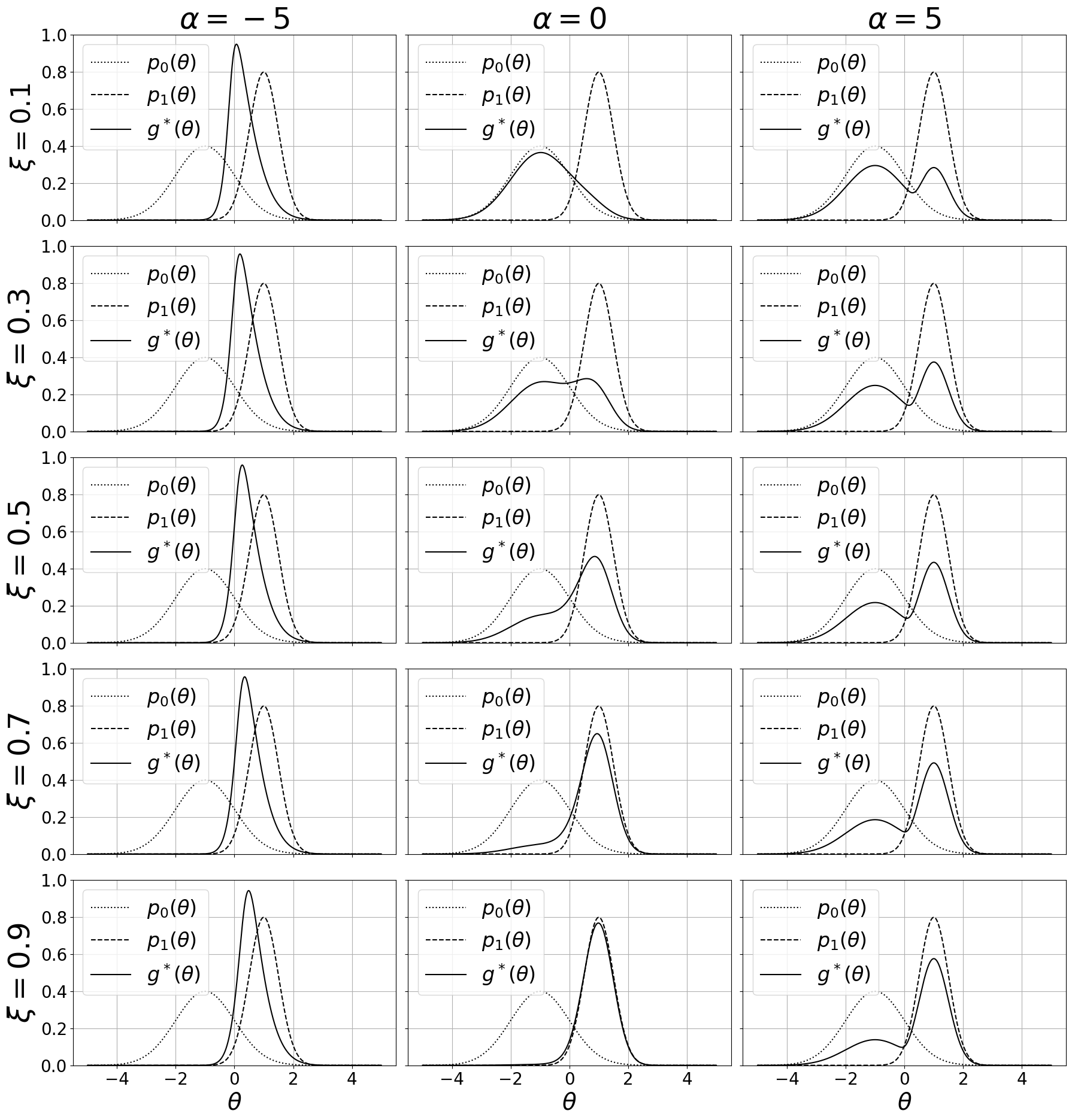}
    \caption{Role of the parameter $\alpha$ for the shape of generalized power posterior (Gaussian distributions, $p_0 = \mathcal{N}(1, 1)$ and $p_1 = \mathcal{N}(-1, 0.25)$). \label{fig:shape_of_power_sum_posterior_gaussian}}
\end{figure}

\begin{figure}[H]
    \centering
    \includegraphics[width=0.95\linewidth]{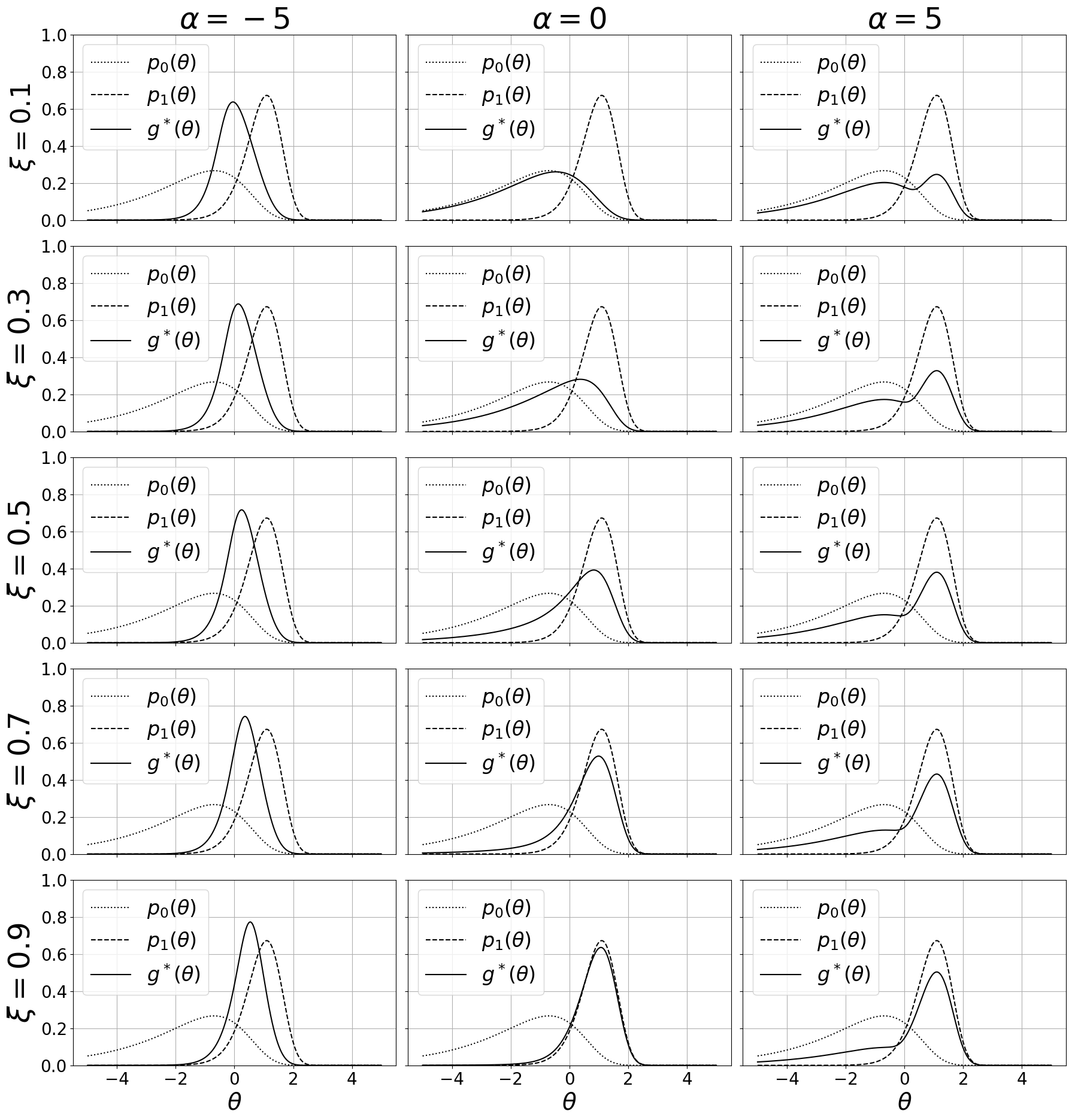}
    \caption{Role of the parameter $\alpha$ for the shape of generalized power posterior (LogGamma distributions, $p_0 = \mathrm{LogGamma}(0.5)$ and $p_1 = \mathrm{LogGamma(3)}$). \label{fig:shape_of_power_sum_posterior_loggamma}}
\end{figure}

\begin{figure}[H]
    \centering
    \includegraphics[width=0.95\linewidth]{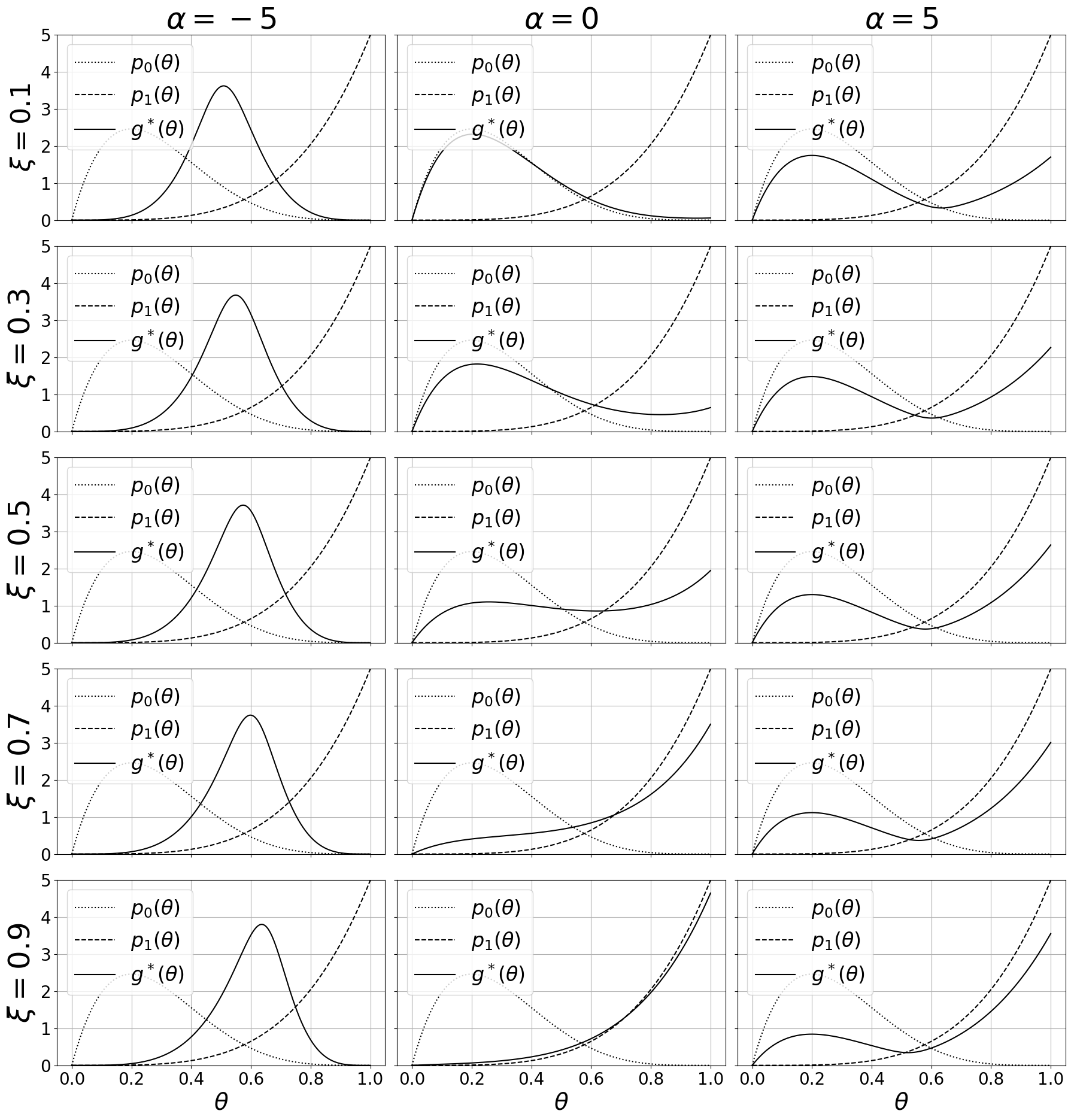}
    \caption{Role of the parameter $\alpha$ for the shape of generalized power posterior (Beta distributions, $p_0 = \mathrm{Beta}(2, 5)$ and $p_1 = \mathrm{Beta}(5, 1)$). \label{fig:shape_of_power_sum_posterior_beta}}
\end{figure}

\begin{assumption}[Unimodality of base distributions]
    Each $\theta \mapsto p_i(\theta)$ is assumed strictly unimodal, with a unique global mode $m_i$.
    That is, there is exactly one $m_i$ such that $p'_i(m_i) = 0$, $p''_i(m_i) < 0$ and $p_i(\theta)$ is increasing on $(-\infty, m_i)$ and decreasing on $(m_i, \infty)$.
\end{assumption}
\begin{assumption}[Mode locations]
    Assume that $m_0 < m_1$.
\end{assumption}
\begin{assumption}[Swapping dominance]
    Assume that there is at least one point $\theta^* \in (m_0, m_1)$ with $p_0(\theta^*) = p_1(\theta^*)$  and $p_0(\theta) > p_1(\theta)$ for $\theta \in (m_0, \theta^*)$, $p_1(\theta) > p_0(\theta)$ for $\theta \in (\theta^*, m_1)$.
    Equivalently, assume that the ratio $p_1(\theta) / p_0(\theta)$ transitions from below $1$ to above $1$ within $(m_0, m_1)$.
\end{assumption}
For the simplicity of the discussion, this subsection assumes a one-dimensional case, but the results obtained can be generalized to multidimensional cases with the same logic.

\begin{lemma}
    \label{lem:derivative_analysis}
    For $z \neq 0$ and two densities $\theta \mapsto p_0(\theta), \theta \mapsto p_1(\theta)$, let
    \begin{align*}
        L_z(\theta) \coloneqq \frac{1}{z}\ln\left\{(1 - \xi)p_0(\theta)^z + \xi p_1(\theta)^z \right\}.
    \end{align*}
    Then, its derivative is given as
    \begin{align*}
        L'_z(\theta) = \frac{p'_0(\theta)}{p_0(\theta)} + \frac{\xi R(\theta)^{z - 1} R'(\theta)}{(1 - \xi) + \xi R(\theta)^z},
    \end{align*}
    where $R(\theta) = p_1(\theta) / p_0(\theta)$.
\end{lemma}

\begin{theorem}
    \label{thm:shape_of_power_sum_posterior}
    Let $\theta \mapsto p_0(\theta)$ and $\theta \mapsto p_1(\theta)$ be two strictly unimodal, continuous density functions on $\mathbb{R}$ with distinct global modes $m_0 < m_1$.
    Suppose there is a nonempty interval $(m_0, m_1)$ on which $p_0$ and $p_1$ swap dominance.
    For $\alpha \neq -1$ and $\xi \in (0, 1)$,
    \begin{itemize}
        \item[i)] for large $\alpha$, generalized power posterior $g$ has at least two local maxima, one near $m_0$ and another near $m_1$. In particular, $g$ is multimodal.
        \item[ii)] for small $\alpha$, generalized power posterior $g$ has exactly one global maximum. Hence $g$ is unimodal.
    \end{itemize}
\end{theorem}

Figure~\ref{fig:shape_of_power_sum_posterior} shows the numerical demonstration of the role of $\alpha$ on the shape of the generalized power posterior.
In this illustrative example, $p_0$ and $p_1$ are two Gaussian distributions centered at $1$ and $-1$, and $\xi = 0.5$.
As stated in Theorem~\ref{thm:shape_of_power_sum_posterior}, it can be observed that the parameter $\alpha$ changes the shape of the posterior distribution.


\subsection{Consistency and Higher-order Asymptotic Analysis}
\label{sec:theory:consistency_and_asymptotic_analysis}
Finally, we consider the following consistency and asymptotic variance of the generalized power posterior.

\begin{theorem}[Consistency of generalized power posterior]
    \label{thm:consistency}
    Let the current data $D = \{X_1,\dots,X_n\}$ and $D_0 = \{Y_1,\dots,Y_{n_0}\}$ be i.i.d. from a parametric model $P_\theta$ with true parameter $\theta_0$.
    Assume that
    \begin{itemize}
        \item $\theta_0 \in \mathrm{int}(\Theta)$,
        \item the likelihoods $L(\theta \mid D)$ and $L(\theta \mid D_0)$ are identifiable, strictly positive and continuous in $\theta$, and
        \item the baseline prior $\pi_0(\theta)$ is strictly positive and continuous in a neighborhood of $\theta_0$.
    \end{itemize}
    Then, the generalized power posterior is consistent for $\theta_0$.
    That is, for any $\epsilon > 0$,
    \begin{align*}
        \pi\left(\left\{\theta \in \Theta\ \colon\ \|\theta - \theta_0\| > \epsilon \right\} \mid D, D_0, \xi, \alpha\right) \overset{P_{\theta_0}}{\underset{n,n_0 \to \infty}{\to}} 0.
    \end{align*}
\end{theorem}
\begin{theorem}[Asymptotic variance of generalized power posterior]
    \label{thm:asymptotic_variance}
    Suppose suitable regularity conditions hold so that a Laplace‐type expansion is valid around $\theta_0 \in \Theta$.
    Then,
    \begin{align*}
        \mathrm{Var}(\hat{\theta}_n) = \frac{1}{n}\left[I(\theta_0) + \frac{\xi n_0}{n}I_0(\theta_0)\right]^{-1} + \frac{1}{n^2}\Gamma_{\alpha,\xi} + o(1/n^2),
    \end{align*}
    where $I(\theta_0)$ and $I(\theta_0)$ are the Fisher‐information‐type matrices based on $D$ and $D_0$, and $\Gamma_{\alpha,\xi}$ is the matrix which depends on $\alpha$ and $\xi$.
\end{theorem}
From the above statement, we can see that the leading term does not depend on $\alpha$, while it appears in the next higher-order term.

\subsection{Information Geometry of Generalized Power Posterior}
Information geometry is a powerful tool to understand statistical procedure in terms of differential geometry~\citep{amari2000methods,amari2016information,ay2017information,nielsen2020elementary}.
The core idea of information geometry is to capture statistical concepts as geometric objects on a Riemannian manifold consisting of a set of probability distributions.
This framework, which combines geometry and statistics, has attracted much attention in recent years for its usefulness, especially in understanding statistical and machine learning procedures and deriving new algorithms~\citep{amari1995information,amari1998natural,ay2015information,hino2024geometry,kimura2021generalized,kimura2021alpha,kimura2022information,kimura2024short,kimura2024density,martens2020new,muller2023achieving,murata2004information}.

Assume that the mapping $\bm{\theta} \mapsto p(X; \bm{\theta})$ is one-to-one.
That is, the probability density function is identifiable by its parameters.
Under this assumption, the parameter space $\Theta$ can be regarded as a coordinate system of the manifold consisting of a set of probability distributions, and one can call this $\bm{\theta}$-coordinate system.
Furthermore, it is known that the Fisher information matrix $\bm{G} = (g_{ij})$ is a metric of this manifold.
Let $\mathcal{M}$ denote this manifold and call it a statistical manifold.
Here, $\mathcal{M}$ is a Riemannian manifold, and there is no guarantee that the Euclidean distance serves as a proper distance function.
Indeed, a class of divergence functions plays this role.

On Riemannian manifolds~\citep{lee2018introduction,lang1995differential}, geometric objects and procedures such as curves and transport depend on the metric equipped by the manifold.
Specifically, consider the straight line, called a geodesic on the $\bm{\theta}$-coordinate system, connecting two distributions $p(X; \bm{\theta}_1)$ and $p(X; \bm{\theta}_2)$.
\begin{align}
    \bm{\theta}(t) = (1 - t)\bm{\theta}_1 + t \bm{\theta}_2, \label{eq:e_geodesics}
\end{align}
where $t \in [0, 1]$.
Assume that $\mathcal{M}$ is the manifold of exponential family distributions.
\begin{align*}
    p(X; \bm{\theta}) &\coloneqq \exp\left\{\sum_{i=1}\theta^i h_i(X) + k(X) - \psi(\bm{\theta}) \right\},
\end{align*}
where $h_i(X)$ are functions of $X$ which are linearly independent, $k(X)$ is a function of $X$, $\psi$ corresponds to the normalization factor.
Then, the probability distributions on the geodesic are
\begin{align*}
    p(X; t) = \{p(X; \bm{\theta}(t))\} = \exp\left\{t(\bm{\theta}_2 - \bm{\theta}_1)\cdot X + \bm{\theta_1} X - \psi(t) \right\}
\end{align*}
Hence, a geodesic itself is a one-dimensional exponential family, where $t$ is the natural parameter.
By taking the logarithm,
\begin{align*}
    \ln p(X; t) = (1 - t) \ln p(X; \bm{\theta}_1) + t \ln p(X; \bm{\theta}_2) - \psi(t).
\end{align*}
Therefore, a geodesic in Eq.~\eqref{eq:e_geodesics} consists of a linear interpolation of the two probability distributions in the logarithmic scale, and it is called an $e$-geodesic, where $e$ stands for exponential.
Using this notation, the $\bm{\theta}$-coordinate system is called $e$-coordinate system.
Note that a submanifold which is defined by linear constraints in $\bm{\theta}$ is said to be $e$-flat.

One of the most interesting concepts in Riemannian manifolds is coordinate transformation.
As with the geodesics above, local geometric objects are dependent on the coordinate system.
That is, considering that the discussion of statistical procedures is carried forward on Riemannian manifolds, the interpretation may change depending on the choice of coordinate system.
In particular, statistical manifolds are known to have a dual coordinate system corresponding to a certain coordinate system.
Let $\bm{\eta}$ be the dual coordinate system of $\bm{\theta}$.
The dual geodesic connecting two distributions specified $\bm{\eta}_1$ and $\bm{\eta}_2$ is given as
\begin{align*}
    \bm{\eta}(t) = (1 - t) \bm{\eta}_1 + t \bm{\eta}_2.
\end{align*}
It is known that this geodesic can be written as
\begin{align*}
    p(X; t) = (1 - t) p_1(X; \bm{\eta}_1) + t p_2(X; \bm{\eta}),
\end{align*}
which is a mixture of two distributions.
This geodesic is called an $m$-geodesic, where $m$ stands for the mixture.

Consider more general coordinate systems.
Specifically, in the literature of information geometry, the $\alpha$-coordinate system is known as one of the most important classes of coordinate systems.
The $\alpha$-coordinate system given as the coordinate transformation as $\bm{m}^{(\alpha)} = p(X; \bm{\theta})^{\frac{1 - \alpha}{2}}$.
Then, the probability distributions on the $\alpha$-geodesic is given as
\begin{align*}
    p(X; t) = C(t)\left\{(1 - t) p(X; \bm{\theta}_1)^{\frac{1 - \alpha}{2}} + t p(X; \bm{\theta}_2)^{\frac{1 - \alpha}{2}} \right\},
\end{align*}
where $C(t)$ is the normalization factor.
In addition, setting $\alpha' = -\alpha$ yields the dual coordinate system and dual geodesic.
That is, the dual $\alpha$-geodesic is given as
\begin{align*}
    p(X; t) = C(t)\left\{(1 - t) p(X; \bm{\theta}_1)^{\frac{1 + \alpha}{2}} + t p(X; \bm{\theta}_2)^{\frac{1 + \alpha}{2}} \right\}.
\end{align*}
This shows that the generalized power posterior can be identified with the $\alpha$-geodesics on the statistical manifold.
One can see that $\alpha \to \pm 1$ yields $e$- and $m$-geodesics, respectively.
\begin{remark}[Information geometry of generalized power posterior]
    The optimal posterior under Amari's $\alpha$-divergence, which is the generalization of the posterior corresponding to the power prior, can be identified with $\alpha$-geodesics on the statistical manifold $\mathcal{M}$.
\end{remark}
On the $\alpha$-coordinate systems, it is known that many geometrical procedures such as $\alpha$-Pythagorean theorem and $\alpha$-projection theorem can be justified.

\section{Application}
\label{sec:application}
We consider the Bayesian survival analysis for Eastern Cooperative Oncology Group (ECOG) phase III clinical trials, labelled E1684 and E1690.
Let $D = \{t_i, \delta_i, g_i\}^n_{i=1}$ be the current trial data (E1690), where $t_i$ is the observed time-to-event for subject, $\delta_i$ is the event indicator (1 if event occurred, 0 if censored), and $g_i$ indicates the treatment group such that $\text{0 = observation}$ and $\text{1 = interferon}$.
The trial of E1684 is a two-arm clinical trial comparing high dose interferon with an observation arm, and $n_0 = 286$ patients were enrolled in the study. 
Let $D_i = \{t^{\text{hist}}_j, \delta^{\text{hist}}_j\}^{n_0}_{j=1}$ be the historical data (E1684) assumed here for the observation arm as $g = 0$.
The findings from this study indicated that interferon alpha-2b significantly improves relapse-free survival and overall survival, prompting the US Food and Drug Administration to approve this regimen as an adjuvant treatment for patients with high-risk melanoma.
In this context, relapse-free survival refers to the duration from randomization to either tumor progression or death, whichever occurs first, while survival is measured as the time from randomization to death.
The findings of the E1684 trial were published by \citet{kirkwood1996interferon}.
This regimen is commonly utilized as adjuvant therapy for patients with high-risk melanoma and serves as the benchmark for assessing alternative treatments, including vaccines, in ongoing US Cooperative Group trials.
The ECOG trial E1690 was a phase III clinical study with three treatment groups: high-dose interferon, low-dose interferon, and an observation group, and this study had $n = 427$ patients on the high dose interferon arm and observation arm.
In our analysis, we use only the data from these two arms of the E1690 trial for compatibility with E1684 trial.

\begin{figure}[H]
    \centering
    \includegraphics[width=0.69\linewidth]{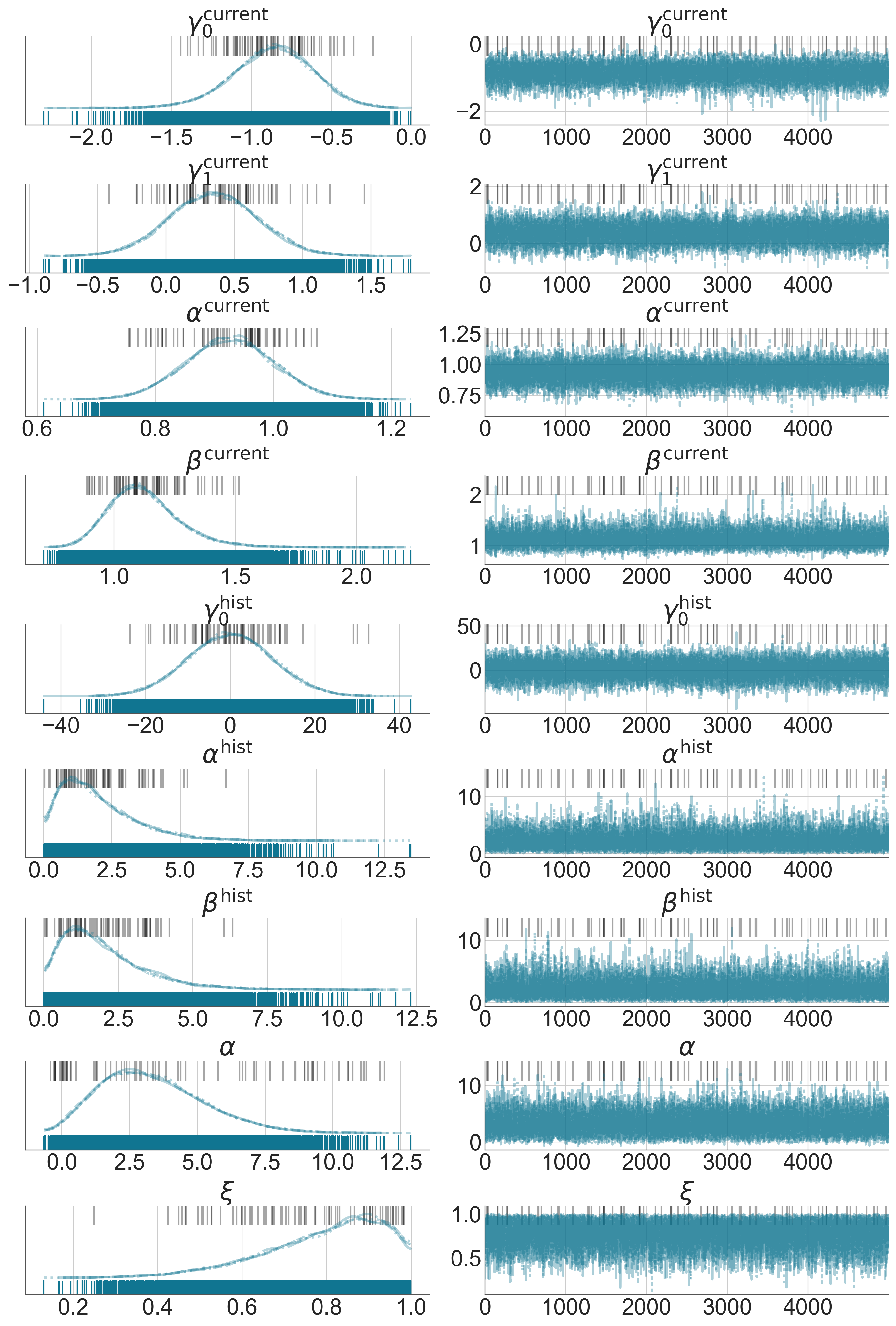}
    \caption{An example of MCMC trace for the model, with $(\mu_\alpha = 0, \sigma_\alpha = 3)$, $(\alpha_\xi = 6, \beta_\xi = 1)$.}
    \label{fig:mcmc_trace}
\end{figure}

For subject $i$ in the current trial, we suppose a cure–rate model where the overall survival function is a mixture of a cured fraction and an uncured Weibull survival component.
Let $\pi_i$ be the cure probability defined as
\begin{align*}
    \pi_i = \sigma\left(\gamma_0^{\text{current}} + \gamma_1^{\text{current}}\cdot g_i\right) = \frac{1}{1 + \exp\left[-(\gamma_0^{\text{current}} + \gamma_1^{\text{current}} \cdot g_i)\right]},
\end{align*}
where $\gamma_0^{\text{current}}$ and $\gamma_1^{\text{current}}$ are regression parameters.
For the uncured subjects with probability $1 - \pi_i$, the survival time follows a Weibull distribution with shape $\alpha^{\text{current}}$ and scale $\beta^{\text{current}}$.
Its probability density function is
\begin{align*}
    f(t_i \mid \alpha^{\text{current}}, \beta^{\text{current}}) = \frac{\alpha^{\text{current}}}{\beta^{\text{current}}}\left(\frac{t_i}{\beta^{\text{current}}}\right)^{\alpha^{\text{current}} - 1}\exp\left[-\left(\frac{t_i}{\beta^{\text{current}}}\right)^{\alpha^{\text{current}}}\right],
\end{align*}
and the corresponding survival function is
\begin{align*}
    S(t_i \mid \alpha^{\text{current}}, \beta^{\text{current}}) = \exp\left[-\left(\frac{t_i}{\beta^{\text{current}}}\right)^{\alpha^{\text{current}}}\right].
\end{align*}
The likelihood contribution is then defined as a mixture:
\begin{align*}
    L_i = \left[(1 - \pi_i)f(t_i \mid \alpha^{\text{current}}, \beta^{\text{current}})\right]^{\delta_i}\left[\pi_i + (1 - \pi_i)S(t_i \mid \alpha^{\text{current}}, \beta^{\text{current}})\right]^{1 - \delta_i}.
\end{align*}
For historical subjects (controls only), we use a separate set of parameters.
Let the historical cure probability be defined by
\begin{align*}
    \pi^{\text{hist}} = \sigma\left(\gamma_0^{\text{hist}}\right) = \frac{1}{1 + \exp\left[-\gamma_0^{\text{hist}}\right]},
\end{align*}
and let the uncured survival be modeled with (potentially different) Weibull parameters $\alpha^{\text{hist}}$ and $\beta^{\text{hist}}$.
Then the density and survival functions are defined as
\begin{align*}
    f^{\text{hist}}(t \mid \alpha^{\text{hist}}, \beta^{\text{hist}}) &= \frac{\alpha^{\text{hist}}}{\beta^{\text{hist}}}\left(\frac{t}{\beta^{\text{hist}}}\right)^{\alpha^{\text{hist}} - 1}\exp\left[-\left(\frac{t}{\beta^{\text{hist}}}\right)^{\alpha^{\text{hist}}}\right], \\
    S^{\text{hist}}(t \mid \alpha^{\text{hist}}, \beta^{\text{hist}}) &= \exp\left[-\left(\frac{t}{\beta^{\text{hist}}}\right)^{\alpha^{\text{hist}}}\right].
\end{align*}
Thus, for historical subject $j$, let the likelihood be $L^{\text{hist}}_j$ and
\begin{align*}
    \ell^{\text{hist}} = \sum_j \ln L_j^{\text{hist}}.
\end{align*}
The full posterior for the model combining current and historical data is then proportional to
\begin{align*}
    \pi(\Theta \mid D, D_0) \propto \left[\prod_{i}L_i\right] \times \left\{(1 - \xi) + \xi \exp\left(z\cdot \ell^{\text{hist}}\right)\right\}^{1/z}\times \pi_0(\Theta),
\end{align*}
where $\Theta$ collects all parameters:
\begin{align*}
    \Theta = \{\gamma_0^{\text{current}}, \gamma_1^{\text{current}}, \alpha^{\text{current}}, \beta^{\text{current}}, \gamma_0^{\text{hist}}, \alpha^{\text{hist}}, \beta^{\text{hist}}, \xi, \alpha\}.
\end{align*}
For a subject with treatment $g$ (0 for control, 1 for treatment), the instantaneous hazard at time $t$ is computed as
\begin{align*}
    h(t \mid g) = \frac{(1 - \pi(g)) f(t \mid \alpha^{\text{current}}, \beta^{\text{current}})}{\pi(g) + (1 - \pi(g)) S(t \mid \alpha^{\text{current}}, \beta^{\text{current}})}.
\end{align*}
The hazard ratio comparing treatment to control at $t_0$ is
\begin{align*}
    \text{HR}(t_0) = \frac{h(t_0 \mid g = 1)}{h(t_0 \mid g = 0)}.
\end{align*}
We set
\begin{align*}
    & \gamma_0^{\text{current}} \sim \mathcal{N}\left(0, 10 \right),  \quad\gamma_1^{\text{current}} \sim \mathcal{N}\left(0, 10 \right), \\
    & \alpha^{\text{current}} \sim \text{Gamma}(2, 1), \quad  \beta^{\text{current}} \sim \text{Gamma}(2, 1), \\
    & \gamma_0^{\text{hist}} \sim \mathcal{N}\left(0, 10 \right), \\
    & \alpha^{\text{hist}} \sim \text{Gamma}(2, 1), \quad \beta^{\text{hist}} \sim \text{Gamma}(2, 1), \\
    & \alpha \sim \mathcal{N}(\mu_\alpha, \sigma_\alpha), \quad \xi \sim \text{Beta}(\alpha_\xi, \beta_\xi),
\end{align*}
and construct $D$ of sample size $n = 100$ by random sampling from the E1690 trial, and $t_0$ to be median follow-up time.
We use Markov Chain Monte Carlo (MCMC) sampling with 50,000 iterations and 10,000 burn-in steps to estimate posterior distributions of the parameters, with 95\% acceptance rate.
Figure~\ref{fig:mcmc_trace} is an example of MCMC trace plot of last 5,000 sample with specific parameters.
The convergence of the Gibbs
sampler was checked using several diagnostic procedures as recommended by \citet{cowles1996markov} including $\hat{R}$ and ESS.
The computing time required to fit the model is approximately 20 min on an Apple M1 Max processor with 64 GB of RAM.

For the model assessment, we consider the concordance index (C-index) which is a measure of the predictive accuracy of a survival model.
It quantifies how well the risk scores of the model agree with the actual ordering of the observed survival times.
Let $r_i$ be the risk score predicted by the model defined as
\begin{align*}
    r_i = \pi_i + (1 - \pi_i) \times \exp\left[-\left(\frac{t_i}{\beta^{\text{current}}}\right)^{\alpha^{\text{current}}}\right].
\end{align*}
Let $\mathcal{P}$ be the set of all comparable pairs.
For each comparable pair $(i, j) \in \mathcal{P}$, define an indicator function as
\begin{align*}
    C_{ij} = \begin{cases}
    1, & \text{if $r_i > r_j$ (correct ordering)}, \\
    0.5 & \text{if $r_i = r_j$ (tie)}, \\
    0 & \text{if $r_i < r_j$ (incorrect ordering)}.
    \end{cases}
\end{align*}
The C-index is the proportion of all comparable pairs that are concordant,
\begin{align*}
    \text{C-index} = \frac{1}{|\mathcal{P}|}\sum_{(i, j) \in \mathcal{P}} C_{ij}.
\end{align*}
$\text{C-index} = 1$ means the perfect prediction, that is, all comparable pairs are correctly ordered, $\text{C-index} = 0.5$ means no better than random chance, and $\text{C-index} < 0.5$ means predictions are in reverse order relative to the observed outcomes.

\begin{table}[t]
    \centering
    \caption{Experimental results for the hierarchical model with generalized power posterior.}
    \begin{tabular}{c|ccccc}
        \toprule
          & $\alpha$ & $\xi$ & HR & HPD Interval & C-index \\
         \midrule
         $\xi = 0$ a.s. & - & $0$ & $0.7155$ & $(0.3461, 1.2513)$ & $0.9624$ \\
         $\xi = 1$ a.s. & - & $1$ & $0.6578$ & $(0.3028, 1.0678)$ & $0.9452$ \\
         \hline
         $\alpha = 1$ a.s., $\xi \sim \text{Beta}(0.5, 0.5)$ & $1$ & $0.2523$ & $0.7044$ & $(0.4100, 1.0582)$ & $0.9619$ \\
         $\alpha = 1$ a.s., $\xi \sim \text{Beta}(2, 2)$ & $1$ & $0.3983$ & $0.7077$ & $(0.4033, 1.0589)$ & $0.9673$ \\
         $\alpha = 1$ a.s., $\xi \sim \text{Beta}(6, 1)$ & $1$ & $0.7512$ & $0.7040$ & $(0.4118, 1.0586)$ & $0.9617$ \\
         $\alpha = 1$ a.s., $\xi \sim \mathcal{U}(0, 1)$ & $1$ & $0.3335$ & $0.7028$ & $(0.4026, 1.0587)$ & $0.9664$ \\
         \hline
         $\alpha\sim \mathcal{N}(0, 1), \xi\sim\text{Beta}(2, 2)$ & $0.5812$ & $0.3450$ & $0.7290$ & $(0.3980, 1.1520)$ & $0.9691$ \\
         $\alpha\sim\mathcal{N}(0, 3), \xi\sim\text{Beta}(2, 2)$ & $3.0415$ & $0.4002$ & $0.7491$ & $(0.3980, 1.1520)$ & $0.9886$ \\
         $\alpha\sim\mathcal{N}(0, 1), \xi\sim\text{Beta}(6, 1)$ & $0.7928$ & $0.7402$ & $0.7813$ & $(0.4012, 1.1317)$ & $0.9872$ \\
         $\alpha\sim\mathcal{N}(0, 3), \xi\sim\text{Beta}(6, 1)$ & $3.360$ & $0.7949$ & $0.7964$ & $(0.4786, 1.1961)$ & $0.9942$ \\ 
         $\alpha\sim\mathcal{N}(0, 1), \xi\sim\mathcal{U}(0, 1)$ & $0.7899$ & $0.7955$ & $0.7945$ & $(0.4723, 1.1966)$ & $0.9903$ \\
         $\alpha\sim\mathcal{N}(0, 3), \xi\sim\mathcal{U}(0, 1)$ & $3.2350$ & $0.7945$ & $0.7951$ & $(0.4753, 1.1971)$ & $0.9911$ \\
         \bottomrule
    \end{tabular}
    \label{tab:weibull_results}
\end{table}

\begin{figure}[t]
    \centering
    \includegraphics[width=\linewidth]{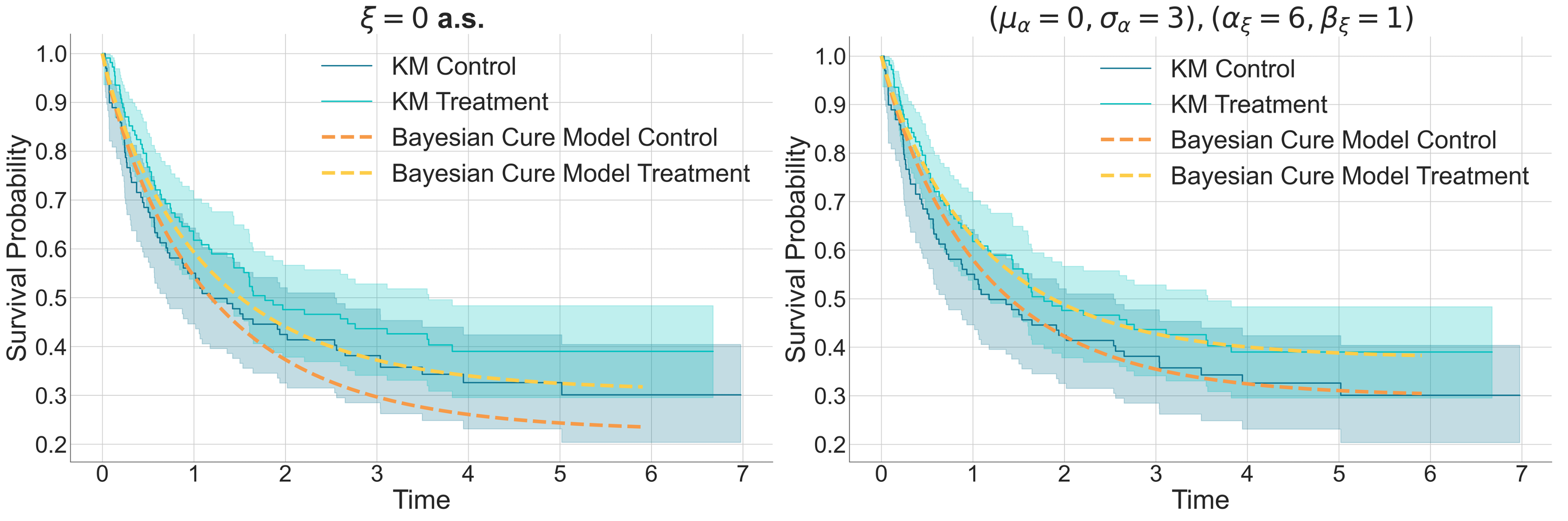}
    \caption{Survival plots for the E1690 trial.}
    \label{fig:km_cure_model}
\end{figure}

\begin{figure}[H]
    \centering
    \includegraphics[width=\linewidth]{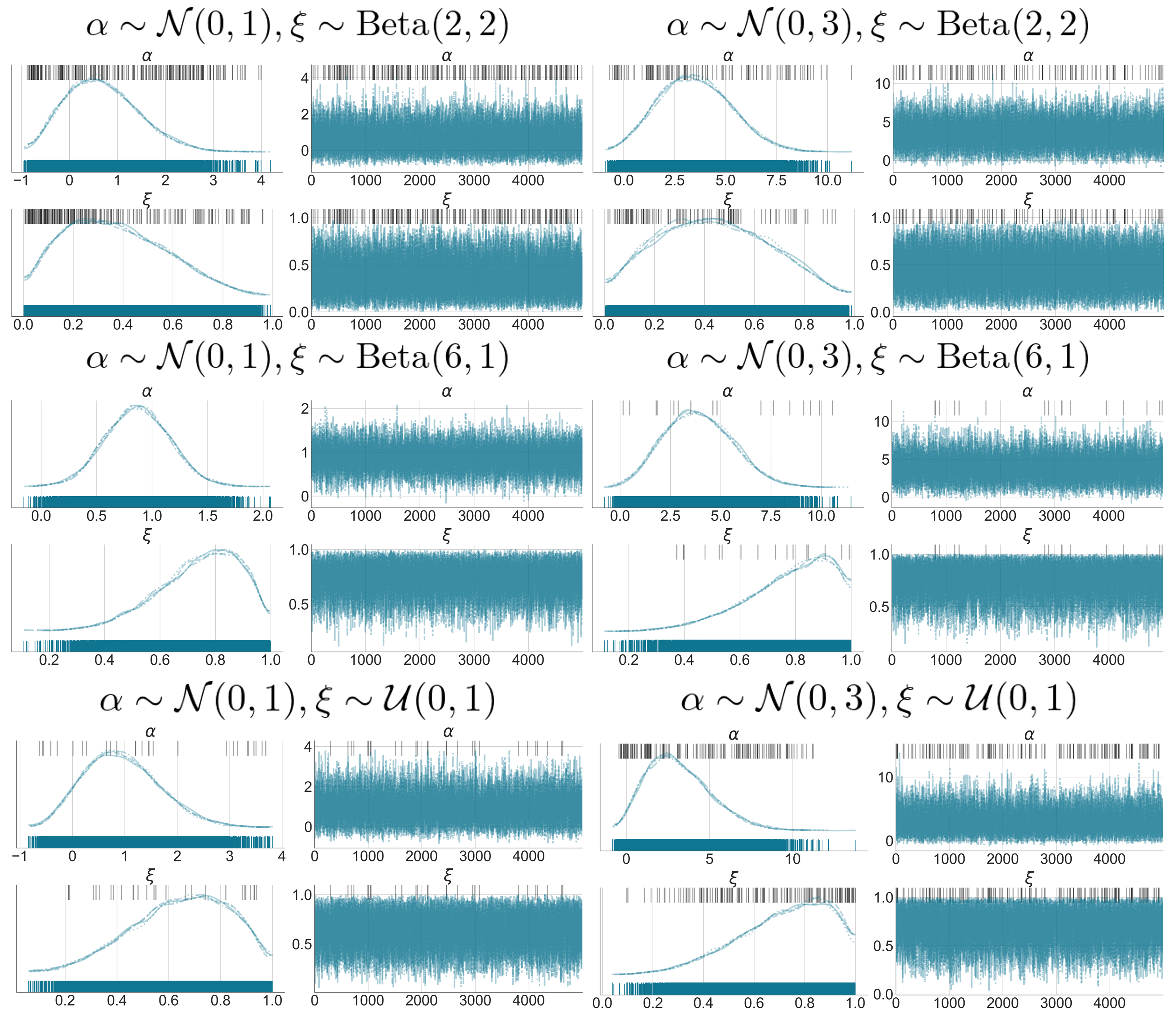}
    \caption{MCMC posteriors for $\alpha$ and $\xi$.}
    \label{fig:mcmc_posterior}
\end{figure}

Experimental results under different hierarchical prior configurations for $(\mu_\alpha, \sigma_\alpha)$ and $(\alpha_\xi, \beta_\xi)$ are summarized in Table~\ref{tab:weibull_results}.
The scenarios of no historical borrowing ($\xi = 0$) and full historical borrowing ($\xi = 1$) are examined as baseline cases.
The model without historical borrowing ($\xi=0$) achieved a hazard ratio (HR) of 0.7155, with a high concordance index (C-index) of 0.9624, reflecting strong predictive performance.
When the model fully relied on historical data ($\xi = 1$), the hazard ratio decreased to 0.6078, and the C-index slightly reduced to 0.9452, indicating some loss in predictive accuracy possibly due to discrepancies between historical and current data.
Adaptive hierarchical borrowing yielded improved predictive accuracy and stable hazard ratio estimates.
Specifically, moderate hierarchical borrowing ($(\mu_\alpha=0, \sigma_\alpha=1), (\alpha_\xi=2, \beta_\xi=2)$ and $(\mu_\alpha=0, \sigma_\alpha=3), (\alpha_\xi=2, \beta_\xi=2)$) maintained HR close to the no-borrowing scenario (approximately 0.709) but significantly enhanced predictive accuracy, achieving C-indices of 0.9691 and 0.9886 respectively.
Stronger adaptive borrowing ($(\mu_\alpha=0, \sigma_\alpha=1), (\alpha_\xi=6, \beta_\xi=1)$ and $(\mu_\alpha=0, \sigma_\alpha=3), (\alpha_\xi=6, \beta_\xi=1)$) increased the hazard ratio slightly to approximately 0.85 but improved predictive performance, attaining C-indices of 0.9872 and 0.9942, respectively.
Moreover, the non-informative uniform prior produces a similar posterior distribution, and it suggests that the data are consistent with larger parameter values.
These results suggest that allowing adaptive borrowing through hierarchical priors effectively leverages historical information without imposing excessive influence on the estimated hazard ratios.
In addition, Figure~\ref{fig:km_cure_model} shows superimposed Kaplan-Meier (KM) estimator~\citep{kaplan1958nonparametric} for full current dataset and posterior mean survival curves of Bayesian hierarchical model by treatment arm for the E1690 trial.
Here, the KM estimator is a non-parametric method used to estimate the survival function from observed survival data, especially in the presence of censored observations.
Given a dataset of survival times $t_{(1)} < t_{(2)} < \cdots < t_{(k)}$, where $t_{(j)}$ represents a distinct observed failure time among the $n$ individuals.
Let $d_j$ be the number of failures at time $t_{(j)}$ and $n_j$ be the number of subjects at risk just prior to time $t_{(j)}$.
Then, the KM estimator of the survival function $S(t)$ is defined as
\begin{align*}
    \hat{S}(t) = \prod_{j\colon t_{(j)} \leq t}\left(1 - \frac{d_j}{n_j}\right), \quad t \geq 0.
\end{align*}
Also, the Greenwood's formula~\citep{greenwood1926natural} for variance is given as
\begin{align*}
    \widehat{\mathrm{Var}}\left(\hat{S}(t)\right) = \hat{S}(t)^2\sum_{j \colon t_{(j)} \leq t}\frac{d_j}{n_j(n_j - d_j)}.
\end{align*}
This variance estimate allows constructing confidence intervals around the KM estimate.
In this figure, we show a common 95\% confidence interval:
\begin{align*}
    \hat{S}(t) = \pm z_{0.975}\sqrt{\widehat{\mathrm{Var}}\left(\hat{S}(t)\right)},
\end{align*}
where $z_{0.975} \approx 1.96$ is the 97.5th percentile of the standard normal distribution.
These result demonstrate that the generalized power posterior approach effectively balances the trade-off between borrowing strength and model complexity, making it suitable for robust Bayesian inference in survival analyses involving historical data.
Furthermore, it can be seen that a good estimator can be obtained by properly setting the prior distribution of each parameter (Figure~\ref{fig:mcmc_posterior} shows the posterior distribution corresponding to each prior distribution).

\section{Conclusion}
\label{sec:conclusion}
This study generalizes the posterior distribution corresponding to power prior, which is known to be optimal under KL divergence, under Amari's alpha-divergence, and investigates its various properties.
The derived posterior distribution was found to depend on the generalization parameters in terms of its shape and sensitivity to the prior distribution.
Moreover, through the lens of information geometry, one can see that this posterior can be identified with the $\alpha$-geodesics on the Riemannian manifold.

Such variants of the Bayesian posterior distribution may lead to improved performance when integrated into appropriate algorithms~\citep{erixon2003reliability,hill1968posterior,huang2023improved,kimura2024test,lewis1997estimating}.
For example, one might take advantage of the fact that the properties of the posterior distribution derived in this study are controlled by a generalization parameter and optimize this parameter directly.

There have been several recent derivations of algorithms based on geodesics and projection theorems under the framework of information geometry~\citep{akaho2004pca,carter2011information,sharp2022parameter}.
Combining these findings with the findings of this study is expected to lead to the derivation of novel algorithms.


\bibliographystyle{agsm}
\bibliography{Bibliography}

\clearpage
\appendix
\begin{center}
{\large\bf SUPPLEMENTARY MATERIAL}
\end{center}

\begin{description}

\item[Title:] Online supplement to \enquote{Generalized Power Priors for Improved Bayesian Inference with Historical Data}.

\end{description}

\section{Auxiliary Results}
\label{sec:auxiliary_results}
For two distinct baseline priors $\pi_0$ and $\tilde{\pi}_0$, one can compare two generalized power posteriors $g^*$ and $\tilde{g}^*$.
We justify the sensitivity via bounds on the total variation distance between the posterior distributions.

\begin{lemma}
    \label{lem:bound_total_variation_distance}
    Let $\bm{\theta} \mapsto g^*(\bm{\theta})$ and $\bm{\theta} \mapsto \tilde{g}^*(\bm{\theta})$ be two generalized power posterior.
    Assume that there exists a measurable set $\mathcal{A} \subset \Theta$ such that $\pi_0(\mathcal{A}) = \tilde{\pi}_0(\mathcal{A}) = 1$, and that for all $\theta \in \mathcal{A}$ there exist constants $0 < m_L \leq 1 \leq M_L < \infty$ subject to,
    \begin{align*}
        m_L &\leq L(\bm{\theta} \mid D) \leq M_L, \\
        m_L &\leq L(\bm{\theta} \mid D_0) \leq M_L,
    \end{align*}
    and $0 < m_\pi \leq 1 \leq M_\pi$ subject to, for all $\bm{\theta} \in \Theta$,
    \begin{align*}
        m_\pi \leq \pi_0(\bm{\theta}) \leq M_\pi, \\
        m_\pi \leq \tilde{\pi}_0(\bm{\theta}) \leq M_\pi,
    \end{align*}
    where $\pi_0(\bm{\theta})$ and $\tilde{\pi}_0(\bm{\theta})$ are the baseline priors corresponding to $g^*$ and $\tilde{g}^*$.
    Then, there exists a function $\alpha \mapsto K(\alpha)$ which depends on $\alpha \in (-1, +\infty)$ and satisfies
    \begin{align}
        d_{\mathrm{TV}}(g^*, \tilde{g}^*) \leq K(\alpha) d_{\mathrm{TV}}(\pi_0, \tilde{\pi}_0), \label{eq:tv_bound_k_alpha}
    \end{align}
    where $(p, q) \mapsto d_{\mathrm{TV}}(p, q)$ is the total variation distance between two probability densities $p$ and $q$:
    \begin{align*}
        d_{\mathrm{TV}} \coloneqq \frac{1}{2}\int_\Theta \left|p(\bm{\theta}) - q(\bm{\theta}) \right| d\bm{\theta}.
    \end{align*}
\end{lemma}
By analyzing the function $K(\alpha)$ in Lemma~\ref{lem:bound_total_variation_distance}, the following theorem can be obtained.
\begin{theorem}[Global prior sensitivity of generalized power posterior]
    \label{thm:global_prior_sensitivity}
    Suppose that $\alpha \in (-1, +\infty)$.
    Then, the resulting generalized power posterior satisfies Eq.~\eqref{eq:tv_bound_k_alpha}, where $K(\alpha)$ takes it minimum at $\alpha = 1$.
\end{theorem}

\section{Proofs}
\label{sec:proofs}

\subsection{Proofs for Section~\ref{sec:background}}

\begin{proof}[Proof for Theorem~\ref{thm:optimality_under_kl_divergence}]
    From the given definitions, for any $\xi \in [0, 1]$,
    \begin{align*}
        & (1 - \xi) D_{\mathrm{KL}}[g \| p_0] + \xi D_{\mathrm{KL}}[g \| p_1] \\
        &\quad\quad\quad = (1 - \xi)\int_\Theta g(\bm{\theta})\ln\frac{g(\bm{\theta})}{p_0(\bm{\theta})}d\bm{\theta} + \xi \int_\Theta g(\bm{\theta})\ln\frac{g(\bm{\theta})}{p_1(\bm{\theta})}d\bm{\theta} \\
        &\quad\quad\quad = \int_\Theta g(\bm{\theta})\left\{(1 - \xi)\ln\frac{g(\bm{\theta})}{p_0(\bm{\theta})} + \xi \ln\frac{g(\bm{\theta})}{p_1(\bm{\theta})} \right\} d\bm{\theta} \\
        &\quad\quad\quad = \int_\Theta g(\bm{\theta})\left\{(1 - \xi) \ln g(\bm{\theta}) - (1 - \xi) \ln p_0(\bm{\theta}) + \xi \ln g(\bm{\theta}) - \xi \ln p_1(\bm{\theta})\right\} d\bm{\theta} \\
        &\quad\quad\quad = \int g(\bm{\theta})\ln g(\bm{\theta}) d\bm{\theta} - (1 - \xi) \int g(\bm{\theta}) \ln p_0(\bm{\theta})d\bm{\theta} - \xi \int g(\bm{\theta})\ln p_1(\bm{\theta})d\bm{\theta}.
    \end{align*}
    We must have $g(\bm{\theta})$ be a probability distribution.
    That is,
    \begin{align*}
        \int_\Theta g(\bm{\theta})d\bm{\theta} = 1.
    \end{align*}
    Then, we impose this constraint using a Lagrange multiplier $\lambda$.
    Define the Lagrangian $(g, \lambda) \mapsto \mathcal{L}(g, \lambda)$ as
    \begin{align*}
        \mathcal{L}(g, \lambda) &\coloneqq (1 - \xi) D_{\mathrm{KL}}[g \| p_0] + \xi D_{\mathrm{KL}}[g \| p_1] + \lambda \left(\int_\Theta g(\bm{\theta})d\bm{\theta} - 1\right) \\
        &= \int g(\bm{\theta})\ln g(\bm{\theta}) d\bm{\theta} - (1 - \xi) \int g(\bm{\theta}) \ln p_0(\bm{\theta})d\bm{\theta} \\
        &\quad\quad\quad - \xi \int g(\bm{\theta})\ln p_1(\bm{\theta})d\bm{\theta} + \lambda \left(\int_\Theta g(\bm{\theta})d\bm{\theta} - 1\right).
    \end{align*}
    To find the minimum, take the functional derivative of $\mathcal{L}$ with respect to $g(\bm{\theta})$ and set it to zero.
    The variations of each term are as follows.
    \begin{align*}
        \frac{\delta}{\delta g(\bm{\theta})}\left\{\int_\Theta g(\bm{\theta})\ln g(\bm{\theta}) d\bm{\theta} \right\} &= \ln g(\bm{\theta}) + 1, \\
        \frac{\delta}{\delta g(\bm{\theta})}\left\{ - (1 - \xi) \int g(\bm{\theta}) \ln p_0(\bm{\theta})d\bm{\theta} \right\} &= -(1 - \xi) \ln p_0(\bm{\theta}), \\
        \frac{\delta}{\delta g(\bm{\theta})}\left\{ - \xi \int g(\bm{\theta})\ln p_1(\bm{\theta})d\bm{\theta} \right\} &= -\xi \ln p_1(\bm{\theta}), \\
        \frac{\delta}{\delta g(\bm{\theta})}\left\{ \lambda \left(\int_\Theta g(\bm{\theta})d\bm{\theta} - 1\right) \right\} &= \lambda.
    \end{align*}
    Putting these together and set this equal to zero.
    \begin{align*}
        \frac{\delta \mathcal{L}(g, \lambda)}{\delta g(\bm{\theta})} &= \ln g(\bm{\theta}) + 1 -(1 - \xi) \ln p_0(\bm{\theta}) -\xi \ln p_1(\bm{\theta}) + \lambda \\
        &= \ln g(\bm{\theta}) + 1 + \lambda - \left\{(1 - \xi)\ln p_0(\bm{\theta}) + \xi \ln p_1(\bm{\theta}) \right\} = 0
    \end{align*}
    Then,
    \begin{align*}
        \ln g(\bm{\theta}) &= (1 - \xi)\ln p_0(\bm{\theta}) + \xi \ln p_1(\bm{\theta}) - 1 - \lambda \\
        g(\bm{\theta}) &= \exp\left\{(1 - \xi)\ln p_0(\bm{\theta}) + \xi \ln p_1(\bm{\theta}) - 1 - \lambda \right\} \\
        &= \exp(-1 - \lambda) \cdot p_0(\bm{\theta})^{1 - \xi}p_1(\bm{\theta})^\xi.
    \end{align*}
    Notice that $\exp(-1 - \lambda)$ is just a normalizing constant, and call it $C$.
    \begin{align*}
        g(\bm{\theta}) = C p_0(\bm{\theta})^{1 - \xi}p_1(\bm{\theta})^\xi.
    \end{align*}
    Here, we need $\int_\Theta g(\bm{\theta}) d\bm{\theta} = 1$, then
    \begin{align*}
        \int_\Theta C p_0(\bm{\theta})^{1 - \xi}p_1(\bm{\theta})^\xi d\bm{\theta} &= 1 \\
        C \int_\Theta p_0(\bm{\theta})^{1 - \xi}p_1(\bm{\theta})^\xi d\bm{\theta} &= 1 \\
        C &= \frac{1}{\int_\Theta p_0(\bm{\theta})^{1 - \xi}p_1(\bm{\theta})^\xi d\bm{\theta}}.
    \end{align*}
    In addition,
    \begin{align*}
        p_0(\bm{\theta})^{1 - \xi}p_1(\bm{\theta})^\xi &= \left\{L(\bm{\theta} \mid D)\pi_0(\bm{\theta})\right\}^{1 - \xi}\left\{L(\bm{\theta} \mid D)L(\bm{\theta} \mid D_0)\pi_0(\bm{\theta})\right\}^\xi \\
        &= L(\bm{\theta} \mid D) L(\bm{\theta} \mid D_0)^\xi \pi_0(\bm{\theta}).
    \end{align*}
    Thus,
    \begin{align*}
        g(\bm{\theta}) = \frac{L(\bm{\theta} \mid D) L(\bm{\theta} \mid D_0)^\xi \pi_0(\bm{\theta})}{\int_\Theta L(\bm{\theta'} \mid D) L(\bm{\theta'} \mid D_0)^\xi \pi_0(\bm{\theta'})d\bm{\theta}'}.
    \end{align*}
    That is, $g(\bm{\theta}) \propto L(\bm{\theta} \mid D)\pi(\bm{\theta}, D, D_0, \xi)$ minimizes the linear combination of KL divergences.
\end{proof}

\clearpage

\subsection{Proofs for Section~\ref{sec:auxiliary_results}}

\begin{proof}[Proof for Lemma~\ref{lem:bound_total_variation_distance}]
    We aim to bound the total variation distance between the two generalized power posterior $g^*$ and $\tilde{g}^*$ in terms of the total variation distance between the baseline priors $\pi_0$ and $\tilde{\pi}_0$.
    Let 
    \begin{align*}
        g^* &= \frac{\left\{(1 - \xi)p_0(\bm{\theta})^z + \xi p_1(\bm{\theta})^z\right\}^{\frac{1}{z}}}{C}, \\
        \tilde{g}^* &= \frac{\left\{(1 - \xi)\tilde{p}_0(\bm{\theta})^z + \xi \tilde{p}_1(\bm{\theta})^z\right\}^{\frac{1}{z}}}{\tilde{C}}, \\
        p_0 &= L(\bm{\theta} \mid D) \pi_0(\bm{\theta}), \\
        \tilde{p}_0 &= L(\bm{\theta} \mid D) \tilde{\pi}_0(\bm{\theta}), \\
        p_1 &= L(\bm{\theta} \mid D) L(\bm{\theta} \mid D_0) \pi_0(\bm{\theta}), \\
        \tilde{p}_1 &= L(\bm{\theta} \mid D) L(\bm{\theta} \mid D_0) \tilde{\pi}_0(\bm{\theta}).
    \end{align*}
    where $z = (1 + \alpha) / 2$ and $C, \tilde{C}$ are normalizing factors.

    First, let
    \begin{align*}
        F(\bm{\theta}) &\coloneqq \left\{(1 - \xi) p_0(\bm{\theta}) + \xi p_1(\bm{\theta}) \right\}^{1 / z}, \\
        \tilde{F}(\bm{\theta}) &\coloneqq \left\{(1 - \xi) \tilde{p}_0(\bm{\theta}) + \xi \tilde{p}_1(\bm{\theta}) \right\}^{1 / z},
    \end{align*}
    and
    \begin{align*}
        \left|g^*(\bm{\theta}) - \tilde{g}^*(\bm{\theta})\right| &= \left|\frac{F(\bm{\theta})}{C} - \frac{\tilde{F}(\bm{\theta})}{\tilde{C}}\right| \\
        &= \left| \left(\frac{F(\bm{\theta})}{C} - \frac{\tilde{F}(\bm{\theta})}{C}\right) + \left(\frac{\tilde{F}(\bm{\theta})}{C} - \frac{\tilde{F}(\bm{\theta})}{\tilde{C}}\right)\right| \\
        &= \left| \frac{F(\bm{\theta}) - \tilde{F}(\bm{\theta})}{C} + \tilde{F}(\bm{\theta})\left(\frac{1}{C} - \frac{1}{\tilde{C}}\right) \right| \\
        &\leq \left| \frac{F(\bm{\theta}) - \tilde{F}(\bm{\theta})}{C} \right| + \left| \tilde{F}(\bm{\theta}) \right| \left| \frac{1}{C} - \frac{1}{\tilde{C}} \right| \\
        &= \frac{\left| F(\bm{\theta}) - \tilde{F}(\bm{\theta}) \right|}{C} + \frac{|\tilde{F}(\bm{\theta})| \cdot |C - \tilde{C}|}{C \tilde{C}}.
    \end{align*}
    To bound $\left|F(\bm{\theta}) - \tilde{F}(\bm{\theta})\right|$, we employ the mean value theorem for $f(x) = x^{t}$, $t > 0$.
    That is, for positive $a, b > 0$ and $c$ between $a$ and $b$,
    \begin{align*}
        |f(a) - f(b)| &= |f'(c)| \cdot |a - b| \\
        &\leq t c^{t - 1} |a - b|,
    \end{align*}
    and $c^{1 / z - 1} \leq \max(a^{1 / z - 1}, b^{1 / z - 1})$.
    Then,
    \begin{align*}
        \left|a^{t} - b^{t}\right| \leq t \max\left(a^{t - 1}, b^{t - 1}\right) |a - b|,
    \end{align*}
    and substitute $t = 1/z$ gives
    \begin{align*}
        \left|a^{1/z} - b^{1/z}\right| \leq \frac{1}{z} \max\left(a^{1/z - 1}, b^{1/z - 1}\right) |a - b|.
    \end{align*}
    Let
    \begin{align*}
        H(\bm{\theta}) &\coloneqq (1 - \xi) p_0(\bm{\theta}) + \xi p_1(\bm{\theta}), \\
        \tilde{H}(\bm{\theta}) &\coloneqq(1 - \xi) \tilde{p}_0(\bm{\theta}) + \xi \tilde{p}_1(\bm{\theta}).
    \end{align*}
    Given assumptions,
    \begin{align*}
        H(\bm{\theta}), \tilde{H}(\bm{\theta}) &\geq (1 - \xi)m_L^z m_\pi^z + \xi m_L^{2z}m_\pi^z \\
        &= m_L^z m_\pi^z((1 - \xi) + \xi m_L^z) > 0.
    \end{align*}
    Thus, applying the mean value theorem,
    \begin{align*}
        |F(\bm{\theta}) - \tilde{F}(\bm{\theta})| &= \left|H(\bm{\theta})^{1 / z} - \tilde{H}(\bm{\theta})^{1 / z} \right| \\
        &\leq \frac{1}{z}\max\left(H(\bm{\theta})^{1/z - 1}, \tilde{H}(\bm{\theta})^{1/z - 1}\right)|H(\bm{\theta}) - \tilde{H}(\bm{\theta})|.
    \end{align*}
    Given that $H(\bm{\theta})$ and $\tilde{H}(\bm{\theta})$ are bounded above by
    \begin{align*}
        H(\bm{\theta}), \tilde{H}(\bm{\theta}) &\leq (1 - \xi) M_L^z M_\pi^z + \xi M_L^{2z}M_\pi^z \\
        &= M_L^z M_\pi^z \left((1 - \xi) + \xi M_L^z \right) \eqqcolon M_H,
    \end{align*}
    and below by
    \begin{align*}
        m_H \coloneqq m_L^z m_\pi^z((1 - \xi) + \xi m_L^z),
    \end{align*}
    we have
    \begin{align*}
        \max\left(H(\bm{\theta})^{1/z - 1}, \tilde{H}(\bm{\theta})^{1/z - 1}\right) &\leq \max\left(m_H^{1/z - 1}, M_H^{1/z - 1}\right) \\
        &= M_H^{1/z - 1} \mathbbm{1}(z \leq 1) + m_H^{1/z - 1}\mathbbm{1}(z > 1) \eqqcolon M^*_H.
    \end{align*}
    Thus,
    \begin{align*}
        |F(\bm{\theta}) - \tilde{F}(\bm{\theta})| &\leq \frac{1}{z} M_H^* \left|H(\bm{\theta}) - \tilde{H}(\bm{\theta})\right| \\
        &= \frac{1}{z}\left\{M_H^{1/z - 1} \mathbbm{1}(z \leq 1) + m_H^{1/z - 1}\mathbbm{1}(z > 1)\right\} \left|H(\bm{\theta}) - \tilde{H}(\bm{\theta})\right|.
    \end{align*}
    Here,
    \begin{align*}
        \left|H(\bm{\theta}) - \tilde{H}(\bm{\theta})\right| &\leq (1 - \xi) \left| p_0(\bm{\theta})^z - \tilde{p}_0(\bm{\theta})^z \right| + \xi \left| p_1(\bm{\theta})^z - \tilde{p}_1(\bm{\theta})^z \right|, \\
        \left|p_0(\bm{\theta})^z - \tilde{p}(\bm{\theta})^z \right| &\leq z\cdot\max\left(p_0(\bm{\theta})^{z - 1}, \tilde{p}_0(\bm{\theta})^{z - 1}\right)|p_0(\bm{\theta}) - \tilde{p}_0(\bm{\theta})| \\
        &= z \cdot M_p^{z - 1} |p_0(\bm{\theta}) - \tilde{p}_0(\bm{\theta})| \\
        \left| p_1(\bm{\theta})^z - \tilde{p}_1(\bm{\theta})^z \right| &\leq z \cdot (M_p M_L)^{z - 1} \left|p_1(\bm{\theta}) - \tilde{p}_1(\bm{\theta})\right|,
    \end{align*}
    where $M_p = \max(p_0(\bm{\theta})^{z - 1}, \tilde{p}_0(\bm{\theta}^{z - 1}))$.
    Given that $p_1(\bm{\theta}) = L(\bm{\theta} \mid D_0)p_0(\bm{\theta})$ and $\tilde{p}_1(\bm{\theta}) = L(\bm{\theta} \mid D_0)\tilde{p}_0(\bm{\theta})$,
    \begin{align*}
        \left|p_1(\bm{\theta}) - \tilde{p}_1(\bm{\theta})\right| &= \left|L(\bm{\theta} \mid D_0)p_0(\bm{\theta}) - L(\bm{\theta} \mid D_0)\tilde{p}_0(\bm{\theta})\right| \\
        &= L(\bm{\theta} \mid D_0) \left|p_0(\bm{\theta}) - \tilde{p}_0(\bm{\theta})\right| \\
        &\leq M_L \left|p_0(\bm{\theta}) - \tilde{p}_0(\bm{\theta})\right|.
    \end{align*}
    Substituiting back,
    \begin{align*}
        \left| H(\bm{\theta}) - \tilde{H}(\bm{\theta}) \right| &\leq z \cdot \left\{(1 - \xi)M_p^{z - 1} + \xi (M_p M_L)^{z - 1} M_L \right\} |p_0(\bm{\theta}) - \tilde{p}_0(\bm{\theta})| \\
        &= z C_1 |p_0(\bm{\theta}) - \tilde{p}_0(\bm{\theta})|,
    \end{align*}
    where $C_1 \coloneqq \left\{(1 - \xi)M_p^{z - 1} + \xi (M_p M_L)^{z - 1} M_L \right\}$.
    Then,
    \begin{align}
        \left|F(\bm{\theta}) - \tilde{F}(\bm{\theta})\right| &\leq \frac{1}{z}C_2 \cdot z C_1 |p_0(\bm{\theta}) - \tilde{p}_0(\bm{\theta})| \nonumber \\
        &= C_1 C_2 |p_0(\bm{\theta}) - \tilde{p}_0(\bm{\theta})|. \label{eq:bound_for_f_f_tilde}
    \end{align}

    Recall that
    \begin{align*}
        C &= \int_\Theta F(\bm{\theta}) d\bm{\theta} \\
        \tilde{C} &= \int_\Theta \tilde{F}(\bm{\theta}) d\bm{\theta}.
    \end{align*}
    Thus,
    \begin{align}
        |C - \tilde{C}| &= \left| \int_\Theta F(\bm{\theta}) d\bm{\theta} - \int_\Theta \tilde{F}(\bm{\theta})d\bm{\theta} \right| \nonumber \\
        &\leq \int_\Theta \left|F(\bm{\theta}) - \tilde{F}(\bm{\theta})\right| d\bm{\theta} \nonumber \\
        &\leq C_1 C_2 \int_\Theta |p_0(\bm{\theta}) - \tilde{p}_0(\bm{\theta})| d\bm{\theta} \nonumber \\
        &= 2 C_1 C_2 d_{\mathrm{TV}}(p_0, \tilde{p}_0) \nonumber \\
        &\leq 2 C_1 C_2 M_L d_{\mathrm{TV}}(\pi_0, \tilde{\pi}_0). \label{eq:bound_for_c_c_tilde}
    \end{align}

    Also,
    \begin{align}
        |\tilde{F}(\bm{\theta})| &= \left|(1 - \xi) \tilde{p}_0(\bm{\theta})^z + \xi \tilde{p}_1(\bm{\theta})^z\right|^{1/z} \nonumber \\
        &\leq \left\{(1 - \xi)M_p^z + \xi M_p^z M_L^z\right\}^{1/z} \nonumber \\
        &= M_p\left\{(1 - \xi) + \xi M_L^z\right\}^{1/z} \eqqcolon M_F. \label{eq:bound_for_f_tilde}
    \end{align}

    Therefore, combining Eqs.~\eqref{eq:bound_for_f_f_tilde},~\eqref{eq:bound_for_c_c_tilde} and~\eqref{eq:bound_for_f_tilde},
    \begin{align*}
        \left|g^*(\bm{\theta}) - \tilde{g}^*(\bm{\theta})\right| \leq \frac{C_1 C_2 |p_0(\bm{\theta}) - \tilde{p}_0(\bm{\theta})|}{C} + \frac{2 C_1 C_2 M_L M_F  d_{\mathrm{TV}}(\pi_0, \tilde{\pi}_0)}{C\tilde{C}},
    \end{align*}
    and
    \begin{align*}
        d_{\mathrm{TV}}(g^*, \tilde{g}^*) &= \frac{1}{2}\int_\Theta \left|g^*(\bm{\theta}) - \tilde{g}^*(\bm{\theta}) \right| d\bm{\theta} \\
        &\leq \frac{1}{2}\int_\Theta \left\{ \frac{C_1 C_2 |p_0(\bm{\theta}) - \tilde{p}_0(\bm{\theta})|}{C} + \frac{2 C_1 C_2 M_L M_F  d_{\mathrm{TV}}(\pi_0, \tilde{\pi}_0)}{C\tilde{C}} \right\} d\bm{\theta} \\
        &= \frac{1}{2}\left\{\frac{C_1 C_2}{C}\int_\Theta |p_0(\bm{\theta}) - \tilde{p}_0(\bm{\theta})| d\bm{\theta} + \frac{2 C_1 C_2 M_L M_F  d_{\mathrm{TV}}(\pi_0, \tilde{\pi}_0)}{C\tilde{C}} \right\} \\
        &\leq \frac{1}{2}\left\{\frac{C_1 C_2}{C}\cdot 2M_L \cdot d_{\mathrm{TV}}(\pi_0, \tilde{\pi}_0) + \frac{2 C_1 C_2 M_L M_F  d_{\mathrm{TV}}(\pi_0, \tilde{\pi}_0)}{C\tilde{C}} \right\} \\
        &= \frac{C_1 C_2 M_L}{C} \left\{1 + \frac{M_F}{\tilde{C}}\right\} d_{\mathrm{TV}}(\pi_0, \tilde{\pi}_0).
    \end{align*}
    Let $K(\alpha) = \frac{C_1 C_2 M_L}{C} \left\{1 + \frac{M_F}{\tilde{C}}\right\}$, and we have the proof.
\end{proof}

\begin{proof}[Proof for Theorem~\ref{thm:global_prior_sensitivity}]
    For analyzing $K(\alpha)$ in Lemma~\ref{lem:bound_total_variation_distance}, let $z = (1 + \alpha) / 2$ and we write
    \begin{align*}
        K(z) = A(z)B(z)D(z),
    \end{align*}
    where
    \begin{align*}
        A(z) &\coloneqq C_2 = M_H^{1/z - 1}\mathbbm{1}(z \leq 1) + m_H^{1/z - 1}\mathbbm{1}(z > 1), \\
        B(z) &\coloneqq C_1 = (1 - \xi)M_p^{z-1} + \xi (M_p M_L)^{z-1}M_L, \\
        D(z) &\coloneqq \frac{1}{C} + \frac{2 M_L M_P Q(z)}{C\tilde{C}}, \quad Q(z) \coloneqq \left\{(1 - \xi) + \xi M_L^z\right\}^{1/z}.
    \end{align*}

    Because $A(z)$ is piecewise-defined expression for $z \leq 1$ and another for $z > 1$, we can check the sign of $A'(z)$ with each region separately.
    For $z \leq 1$, $A(z) = M_H^{1/z - 1}$.
    Observe that $M_H \geq 1$ implies that $\ln M_H \geq 0$.
    Then,
    \begin{align*}
        A'(z) &= \frac{d}{dz}M_H^{1/z - 1} \\
        &= M_H^{1/z - 1} \ln M_H \frac{d}{dz}\left(\frac{1}{z} - 1\right) \\
        &= M_H^{1/z - 1} \ln M_H\left(-\frac{1}{z^2}\right) \leq 0.
    \end{align*}
    For $z > 1$, $A(z) = m_H^{1/z - 1}$.
    Here, $m_H^{1/z - 1} \leq 1$ implies that $\ln m_H \leq 0$, and
    \begin{align*}
        A'(z) &= \frac{d}{dz}m_H^{1/z - 1} \\
        &= m_H^{1/z - 1} \ln m_H\left(-\frac{1}{z^2}\right) \geq 0.
    \end{align*}

    For $B(z)$,
    \begin{align*}
        B'(z) = (1 - \xi) M_p^{z - 1} \ln M_p + \xi M_L (M_p M_L)^{z - 1} \ln M_p M_L > 0.
    \end{align*}

    For $D(z)$,
    \begin{align*}
        D'(z) = \frac{2M_L M_p}{C\tilde{C}}Q'(z).
    \end{align*}
    Here,
    \begin{align*}
        \ln Q(z) &= \frac{1}{z}\ln\left\{(1 - \xi) + \xi M_L^z\right\}, \\
        \frac{d}{dz} \ln Q(z) &= \frac{d}{dz}\left(\frac{1}{z}\right)\ln\left\{(1 - \xi) + \xi M_L^z \right\} + \frac{1}{z}\frac{\xi M_L^z \ln M_L}{(1 - \xi) + \xi M_L^z} \\
        &= -\frac{1}{z^2}\ln\left\{(1 - \xi) + \xi M_L^z \right\} + \frac{1}{z}\frac{\xi M_L^z \ln M_L}{(1 - \xi) + \xi M_L^z} = \frac{Q'(z)}{Q(z)}, \\
        Q'(z) &= Q(z)\left\{\frac{1}{z}\frac{\xi M_L^z \ln M_L}{(1 - \xi) + \xi M_L^z} - \frac{1}{z^2}\ln\left((1 - \xi) + \xi M_L^z \right)\right\}.
    \end{align*}
    Then,
    \begin{align*}
        D'(z) = \frac{2M_L M_p}{C\tilde{C}} Q(z) \left\{\frac{1}{z}\frac{\xi M_L^z \ln M_L}{(1 - \xi) + \xi M_L^z} - \frac{1}{z^2}\ln\left((1 - \xi) + \xi M_L^z \right)\right\}.
    \end{align*}
    We see that $\frac{2M_L M_p}{C\tilde{C}} > 0$ and $Q(z) > 0$ for all $z > 0$.
    Thus, the sign of $D'(z)$ is determined by the bracket term.

    Let $u \coloneqq (1 - \xi) + \xi M_L^z$, which is positive for all $z, \xi$, and
    \begin{align*}
        h(z) &\coloneqq \frac{1}{z}\frac{\xi M_L^z \ln M_L}{(1 - \xi) + \xi M_L^z} - \frac{1}{z^2}\ln\left((1 - \xi) + \xi M_L^z \right) \\
        &= \frac{\xi M_L^z \ln M_L}{zu} - \frac{1}{z^2}\ln u, \\
        \bar{h}(z) &\coloneqq z \xi M_L^z \ln M_L - u \ln u.
    \end{align*}
    By the product rule,
    \begin{align*}
        \frac{d}{dz}\left(u \ln u \right) &= u'(z) \ln u(z) + u(z)\frac{u'(z)}{u(z)} \\
        &= u'(z) \ln u(z) + u'(z) \\
        &= u'(z) \left(\ln u(z) + 1 \right).
    \end{align*}
    Since $u'(z) = \xi \ln M_L \cdot M_L^z$,
    \begin{align*}
        \frac{d}{dz}\left(u \ln u\right) = \ln M_L \cdot a(z)\left(\ln u(z) + 1\right),
    \end{align*}
    where $a(z) \coloneqq \xi M_L^z$.
    Also,
    \begin{align*}
        \frac{d}{dz}\left(z a(z) \ln M_L \right) &= \ln M_L \cdot \frac{d}{dz}\left(z \cdot a(z)\right) \\
        &= \ln M_L \cdot \left(a(z) + z a'(z)\right) \\
        &= \ln M_L \cdot \left(a(z) + z a(z) \ln M_L \right) \\
        &= a(z) \ln M_L + z a(z) (\ln M_L)^2.
    \end{align*}
    Then,
    \begin{align*}
        \bar{h}'(z) &= \frac{d}{dz}\left(z a(z) \ln M_L \right) - \frac{d}{dz}\left(u(z) \ln u(z) \right) \\
        &= a(z) \ln M_L + z a(z) (\ln M_L)^2 - \ln M_L \cdot a(z) \left(\ln u(z) + 1\right) \\
        &= a(z) \ln M_L \cdot \left(1 + z \ln M_L - \ln u(z) - 1\right) \\
        &= a(z) \ln M_L \cdot \left(z \ln M_L - \ln u(z) \right) \\
        &= a(z) \ln M_L \cdot \left(\ln M_L^z - \ln\left((1 - \xi) + \xi M_L^z \right) \right) \\
        &= a(z) \ln M_L \cdot \ln \left\{\frac{M_L^z}{(1 - \xi) + \xi M_L^z} \right\}
    \end{align*}
    Therefore, we have $D'(z) < 0$ for $z \leq 1$ and $D'(z) > 0$ for $z > 1$.

    Recall that
    \begin{align*}
        K'(z) = A'(z) B(z) D(z) + A(z) B'(z) D(z) + A(z) B(z) D'(z).
    \end{align*}
    For $z \leq 1$, $A'(z), D'(z) < 0$, $A(z), B(z), D(z), B'(z) > 0$, and since $A'(z)$ is order $\frac{\ln M_H}{z^2} M_H^{1/z - 1}$, $B'(z)$ is order $\frac{\ln((1 - \xi) + \cdots)}{z^2}$ and $B'(z)$ is order $(1 - \xi)M_p^{z - 1} \ln M_p$ plus a positive term, $K'(z) < 0$.
    This establishes the decreasing behavior of $K(z)$ in that regime.
    For $z > 1$, $A'(z), B'(z), D'(z) > 0$ and $K'(z) > 0$.
    This shows $K(z)$ is strictly increasing when $z > 1$.
    Putting both cases together, we obtain the proof.
\end{proof}

\subsection{Proofs for Section~\ref{sec:theory}}

\begin{proof}[Proof for Theorem~\ref{thm:robustness_of_alpha_power_sum_posterior}]
    Let us write
    \begin{align*}
        p(\theta) &= g^*(\theta;P_H), \\
        q(\theta) &=  g^*(\theta;P).
    \end{align*}
    Then the total variation distance is
    \begin{align*}
        d_{\mathrm{TV}}(p,q) &= \frac{1}{2}\int\bigl|\,p(\theta)-q(\theta)\bigr|\;d\theta \\
        &= \frac{1}{2}\int q(\theta)\;\left|\frac{p(\theta)}{q(\theta)} - 1\right|\,d\theta \\
        &\leq \frac{1}{2}\left(\sup_{\theta}\;\left|\frac{p(\theta)}{q(\theta)}-1\right|\right)\int q(\theta)\,d\theta.
    \end{align*}
    Since $\int q(\theta)\,d\theta=1$, we reduce the problem to finding $\sup_{\theta}\bigl|\frac{p(\theta)}{q(\theta)}-1\bigr|$.  
    Recall that
    \begin{align*}
        p(\theta) &= \frac{\bigl[(1-\xi)\,p_0(\theta)^z + \xi\,p_1(\theta)^z\bigr]^{\frac1z}}{\displaystyle\int\Bigl[(1-\xi)\,p_0(\theta')^z + \xi\,p_1(\theta')^z\Bigr]^{\frac1z}d\theta'}, \\
        q(\theta) &= \frac{\bigl[(1-\xi)\,p_0^F(\theta)^z + \xi\,p_1^F(\theta)^z\bigr]^{\frac1z}}{\displaystyle\int\Bigl[(1-\xi)\,p_0^F(\theta')^z + \xi\,p_1^F(\theta')^z \Bigr]^{\frac1z} d\theta'}.
    \end{align*}
    Thus
    \begin{align*}
        \frac{p(\theta)}{q(\theta)} = \frac{\bigl[(1-\xi)\,p_0(\theta)^z + \xi\,p_1(\theta)^z\bigr]^{\frac1z}}{\bigl[(1-\xi)\,p_0^F(\theta)^z + \xi\,p_1^F(\theta)^z\bigr]^{\frac1z}} \times \frac{\displaystyle\int\Bigl[(1-\xi)\,p_0^F(\theta')^z + \xi\,p_1^F(\theta')^z \Bigr]^{\frac1z}d\theta'}{\displaystyle \int \Bigl[(1-\xi)\,p_0(\theta')^z + \xi\,p_1(\theta')^z \Bigr]^{\frac1z} d\theta'}.
    \end{align*}
    One then observes that
    \begin{align*}
        \int\Bigl[(1-\xi)\,p_0(\theta')^z + \xi\,p_1(\theta')^z\Bigr]^{\frac1z}d\theta' \geq \int \Bigl[(1-\xi)\,\bigl(p_0^F(\theta')\bigr)^z + \xi\,\bigl(p_1^F(\theta')\bigr)^z\Bigr]^{\frac1z} d\theta',
    \end{align*}
    and
    \begin{align*}
        \frac{p(\theta)}{q(\theta)} &\leq \frac{\bigl[(1-\xi)\,p_0(\theta)^z + \xi\,p_1(\theta)^z\bigr]^{\frac1z}}{ \bigl[(1-\xi)\,p_0^F(\theta)^z + \xi\,p_1^F(\theta)^z\bigr]^{\frac1z}} \\
    &=  \left[\frac{(1-\xi)\,p_0(\theta)^z + \xi\,p_1(\theta)^z}{(1-\xi)\,p_0^F(\theta)^z + \xi\,p_1^F(\theta)^z}\right]^{\frac1z}.
    \end{align*}
    Similarly, considering the reciprocal $q(\theta)/p(\theta)$ yields the matching lower bound.
    Altogether one obtains
    \begin{align*}
        \bigl[\inf_\theta R(\theta)\bigr]^{\frac1z} \leq  \frac{p(\theta)}{q(\theta)} \leq \bigl[\sup_\theta R(\theta)\bigr]^{\frac1z},
    \end{align*}
    where
    \begin{align*}
        R(\theta) = \frac{(1-\xi)\,\bigl[p_0(\theta)\bigr]^z + \xi\,\bigl[p_1(\theta)\bigr]^z}{(1-\xi)\,\bigl[p_0^F(\theta)\bigr]^z + \xi\,\bigl[p_1^F(\theta)\bigr]^z}.
    \end{align*}
    Thus letting
    \begin{align*}
        R_{\min} &= \inf_\theta\,R(\theta), \\
        R_{\max} &= \sup_\theta\,R(\theta),
    \end{align*}
    we can write as
    \begin{align*}
        \frac{p(\theta)}{q(\theta)} \in  \left[\, (R_{\min})^{\frac{1}{z}},\;(R_{\max})^{\frac{1}{z}}\right].
    \end{align*}
    Hence
    \begin{align*}
        \sup_\theta\;\frac{p(\theta)}{q(\theta)} \leq (R_{\max})^{\frac1z}, \quad \inf_\theta\;\frac{p(\theta)}{q(\theta)} \geq (R_{\min})^{\frac1z}.
    \end{align*}
    It follows that
    \begin{align*}
        \sup_{\theta}\;\Bigl|\frac{p(\theta)}{q(\theta)}-1\Bigr| \leq  \max\bigl\{\, (R_{\max})^{1/z} -1,\;\; 1 - (R_{\min})^{1/z} \bigr\},
    \end{align*}
    and in turn one obtains the simpler one‐sided ratio bound $\sup_{\theta}\;\frac{p(\theta)}{q(\theta)} \leq  \Bigl(\frac{R_{\max}}{R_{\min}}\Bigr)^{\!\frac1z}$, which implies 
    \begin{align*}
        \left|\frac{p(\theta)}{q(\theta)}-1\right| \leq \left(\frac{R_{\max}}{R_{\min}}\right)^{\!\frac1z} -1.
    \end{align*}
    Combining the above with
    \begin{align*}
        d_{\mathrm{TV}}(p,q) = \frac{1}{2} \int\bigl|p(\theta)-q(\theta)\bigr|\;d\theta \leq \frac12\sup_{\theta}\,\left|\frac{p(\theta)}{q(\theta)}-1\right|,
    \end{align*}
    one can yield
    \begin{align*}
        d_{\mathrm{TV}}(p,q) \leq \frac12\left[\Bigl(\frac{R_{\max}}{R_{\min}}\Bigr)^{\!\frac1z} -1 \right].
    \end{align*}
    Rewriting $p(\theta)=g^*(\theta;P_H)$ and $q(\theta)=g^*(\theta;P)$ re‐establishes the statement in the theorem,
    \begin{align*}
        d_{\mathrm{TV}}\!\Bigl(g^*(\cdot;P_H),\,g^*(\cdot;P)\Bigr) \leq  \frac12\Bigl[\Bigl(\frac{R_{\max}}{R_{\min}}\Bigr)^{\frac{1}{z}} - 1\Bigr].
    \end{align*}
    
    We have
    \begin{align*}
        p_0(\theta) &= L(\theta)\,\pi_0(\theta), \\
        p_1(\theta) &= \Bigl[L(\theta)\Bigr]^{\frac{n}{n+n_0}}\,L_0(\theta)\,\pi_0(\theta), \\
        p_0^F(\theta) &= L^F(\theta)\,\pi_0(\theta), \\
        p_1^F(\theta) &= \Bigl[L^F(\theta)\Bigr]^{\frac{n}{n+n_0}}\,L_0(\theta)\,\pi_0(\theta).
    \end{align*}
    Define the basic ratio of contaminated to clean likelihood as
    \begin{align*}
        a_0(\theta) = \frac{p_0(\theta)}{p_0^F(\theta)} = \frac{L(\theta)}{L^F(\theta)},
    \end{align*}
    and observe that
    \begin{align*}
        a_1(\theta) = \frac{p_1(\theta)}{p_1^F(\theta)} = \Bigl[a_0(\theta)\Bigr]^{\frac{n}{n+n_0}}.
    \end{align*}
    We also define
    \begin{align*}
        b(\theta) = \frac{p_1^F(\theta)}{p_0^F(\theta)} = \frac{\bigl[L^F(\theta)\bigr]^{\frac{n}{n+n_0}}\,L_0(\theta)\,\pi_0(\theta)}{L^F(\theta)\,\pi_0(\theta)} = \frac{L_0(\theta)}{\bigl[L^F(\theta)\bigr]^{\frac{n_0}{n+n_0}}}.
    \end{align*}
    Then,
    \begin{align*}
        \frac{(1-\xi)\,\bigl[p_0(\theta)\bigr]^z + \xi\,\bigl[p_1(\theta)\bigr]^z}{(1-\xi)\,\bigl[p_0^F(\theta)\bigr]^z + \xi\,\bigl[p_1^F(\theta)\bigr]^z} = \frac{(1-\xi)\,\bigl[a_0(\theta)\bigr]^z + \xi\,\bigl[a_1(\theta)\,b(\theta)\bigr]^z}{(1-\xi) + \xi\,\bigl[b(\theta)\bigr]^z}.
    \end{align*}
    Hence
    \begin{align*}
        R(\theta) = \frac{(1-\xi)\,[a_0(\theta)]^z + \xi\,[b(\theta)\,a_1(\theta)]^z}{(1-\xi) + \xi\,[\,b(\theta)\,]^z}.
    \end{align*}
    Thus to bound $R(\theta)$, it suffices to place bounds on $a_0(\theta)$, $a_1(\theta)$, and $b(\theta)$.
    Recall
    \begin{align*}
        L(\theta) &= \prod_{i=1}^n\Bigl[(1-\epsilon_H)\,f_N(X_i;\theta,\sigma^2) + \epsilon_H\,f_N\bigl(X_i;\theta_H,\sigma^2\bigr)\Bigr], \\
        L^F(\theta) &= \prod_{i=1}^n f_N\bigl(X_i;\theta,\sigma^2\bigr).
    \end{align*}
    Thus
    \begin{align*}
        \frac{L(\theta)}{L^F(\theta)} &= \prod_{i=1}^n\frac{(1-\epsilon_H)\,f_N(X_i;\theta,\sigma^2) + \epsilon_H\,f_N(X_i;\theta_H,\sigma^2)}{f_N(X_i;\theta,\sigma^2)} \\
        &= \prod_{i=1}^n\Bigl[1 - \epsilon_H + \epsilon_H\, \frac{f_N(X_i;\theta_H,\sigma^2)}{f_N(X_i;\theta,\sigma^2)}\Bigr].
    \end{align*}
    Now, since $\Delta_H = |\theta_H-\theta_0|$, the ratio $\frac{f_N(x;\theta_H,\sigma^2)}{f_N(x;\theta,\sigma^2)}$ can be bounded above and below by exponential functions of $\Delta_H^2/\sigma^2$.
    Indeed, for Gaussian densities
    \begin{align*}
        f_N(x;\mu,\sigma^2) = \frac{1}{\sqrt{2\pi\sigma^2}}\,\exp\!\Bigl[-\frac{(x-\mu)^2}{2\sigma^2}\Bigr],
    \end{align*}
    one can show that
    \begin{align*}
        \max_{x,\theta}\;\frac{f_N(x;\theta_H,\sigma^2)}{f_N(x;\theta,\sigma^2)} &= \exp\!\Bigl(\frac{\Delta_H^2}{2\sigma^2}\Bigr), \\
        \min_{x,\theta}\;\frac{f_N(x;\theta_H,\sigma^2)}{f_N(x;\theta,\sigma^2)} &= \exp\!\Bigl(-\frac{\Delta_H^2}{2\sigma^2}\Bigr).
    \end{align*}
    Hence for each factor
    \begin{align*}
        1 - \epsilon_H + \epsilon_H\,\exp\!\Bigl(-\frac{\Delta_H^2}{2\sigma^2}\Bigr) &\leq 1 - \epsilon_H + \epsilon_H\, \frac{f_N(X_i;\theta_H,\sigma^2)}{f_N(X_i;\theta,\sigma^2)} \\
        &\leq 1 - \epsilon_H + \epsilon_H\,\exp\!\Bigl(\frac{\Delta_H^2}{2\sigma^2}\Bigr).
    \end{align*}
    Multiply these inequalities over $i=1,\dots,n$ to conclude
    \begin{align*}
         \bigl[m_0(\Delta_H,\sigma,\epsilon_H)\bigr]^n &\leq a_0(\theta) = \frac{L(\theta)}{L^F(\theta)} \\
         &\leq  \bigl[M_0(\Delta_H,\sigma,\epsilon_H)\bigr]^n,
    \end{align*}
    where we define the constants
    \begin{align*}
        m_0(\Delta_H,\sigma,\epsilon_H) &= 1 - \epsilon_H + \epsilon_H\,\exp\!\Bigl(-\frac{\Delta_H^2}{2\sigma^2}\Bigr), \\
        M_0(\Delta_H,\sigma,\epsilon_H) &=  1 - \epsilon_H + \epsilon_H\,\exp\!\Bigl(\frac{\Delta_H^2}{2\sigma^2}\Bigr).
    \end{align*}
    Note that $m_0\le1\le M_0$ and, as $\epsilon_H\to0$ or $\Delta_H\to0$, both expressions approach to $1$.
    Since
    \begin{align*}
         a_1(\theta) = \Bigl[a_0(\theta)\Bigr]^{\frac{n}{n+n_0}},
    \end{align*}
    it follows immediately that
    \begin{align*}
        \bigl[m_0(\Delta_H,\sigma,\epsilon_H)\bigr]^{\,\frac{n^2}{n+n_0}} \leq a_1(\theta) \leq  \bigl[M_0(\Delta_H,\sigma,\epsilon_H)\bigr]^{\,\frac{n^2}{n+n_0}}.
    \end{align*}
    Recall
    \begin{align*}
        b(\theta) = \frac{p_1^F(\theta)}{p_0^F(\theta)} = \frac{[\,L^F(\theta)\,]^{\frac{n}{n+n_0}}\,L_0(\theta)\,\pi_0(\theta)}{L^F(\theta)\,\pi_0(\theta)} = \frac{L_0(\theta)}{\bigl[L^F(\theta)\bigr]^{\,\frac{n_0}{n+n_0}}}.
    \end{align*}
    Hence $b(\theta)$ reflects how compatible $\theta$ is with the historical likelihood $L_0(\theta)$ relative to the current data likelihood $L^F(\theta)$ but raised to partial powers. 
    A standard argument shows that $\sup_\theta b(\theta)$ and $\inf_\theta b(\theta)$ can each be bounded by exponentials in $(n_0\,\Delta_H^2/\sigma^2)$, plus constants.
    Denote
    \begin{align*}
        \underline{B} \leq b(\theta) \leq \overline{B}, \quad \text{for all }\theta.
    \end{align*}
    Here,
    \begin{align*}
        R(\theta) = \frac{(1-\xi)\,[a_0(\theta)]^z + \xi\,[\,b(\theta)\,a_1(\theta)\bigr]^z}{(1-\xi) + \xi\,[b(\theta)]^z}.
    \end{align*}
    Hence
    \begin{align*}
        \inf_\theta R(\theta) &\geq \inf_\theta \frac{(1-\xi)\,[a_0(\theta)]^z + \xi\,\bigl[b(\theta)\,a_1(\theta)\bigr]^z}{(1-\xi) + \xi\,[b(\theta)]^z} \\
        &\geq \frac{(1-\xi)\,\bigl[\inf_\theta a_0(\theta)\bigr]^z + \xi\,\bigl[\inf_\theta (b(\theta)\,a_1(\theta))\bigr]^z}{(1-\xi) + \xi\,\bigl[\sup_\theta b(\theta)\bigr]^z},
    \end{align*}
    and similarly
    \begin{align*}
        \sup_\theta R(\theta) \leq \frac{(1-\xi)\,\bigl[\sup_\theta a_0(\theta)\bigr]^z + \xi\,\bigl[\sup_\theta (b(\theta)\,a_1(\theta))\bigr]^z}{(1-\xi) + \xi\,\bigl[\inf_\theta b(\theta)\bigr]^z}.
    \end{align*}
    We have already seen
    \begin{align*}
        a_0(\theta) &\in \bigl[m_0(\Delta_H,\sigma,\epsilon_H)^n,\; M_0(\Delta_H,\sigma,\epsilon_H)^n \bigr], \\
        a_1(\theta) &\in \bigl[m_0(\Delta_H,\sigma,\epsilon_H)^{\frac{n^2}{n+n_0}},\; M_0(\Delta_H,\sigma,\epsilon_H)^{\frac{n^2}{n+n_0}}\bigr], \\
        b(\theta) &\in [\,\underline{B},\;\overline{B}\,].
    \end{align*}
    It follows that $b(\theta)\,a_1(\theta)$ also lies in a known interval,
    \begin{align*}
        b(\theta)\,a_1(\theta) \in \Bigl[\underline{B}\,\bigl(m_0(\Delta_H,\sigma,\epsilon_H)\bigr)^{\frac{n^2}{n+n_0}}, \quad \overline{B}\,\bigl(M_0(\Delta_H,\sigma,\epsilon_H)\bigr)^{\frac{n^2}{n+n_0}}\Bigr].
    \end{align*}
    Hence $R_{\min}$ and $R_{\max}$ can each be written entirely in terms of
    \begin{align*}
        \underline{B},\;\overline{B},\; m_0(\Delta_H,\sigma,\epsilon_H),\;M_0(\Delta_H,\sigma,\epsilon_H),\;\xi,\;z=\frac{1+\alpha}{2},\;n,\;n_0.
    \end{align*}
    All these pieces are exponentials or products of known parameters.
    Here,
    \begin{itemize}
        \item As $\epsilon_H\to0$ or $\Delta_H\to0$, we have $L(\theta)=L^F(\theta)$ exactly, so $a_0(\theta)\equiv1$, $a_1(\theta)\equiv1$.  Then $p_0(\theta)\equiv p_0^F(\theta)$ and $p_1(\theta)\equiv p_1^F(\theta)$, hence $R(\theta)\equiv1$ for all $\theta$, forcing $R_{\max}/R_{\min}=1$, so the bound is 0.
        \item As $\xi\to0$ or $\xi\to1$, the mixture collapses and the exponent $z$ no longer influences the mixture (it is effectively $(p_0)^z$ only or $(p_1)^z$ only). Thus the $\alpha$‐dependence (via $z$) vanishes at those endpoints.  The bound itself may remain positive if $\Delta_H\neq0\neq\epsilon_H$, but that positivity no longer depends on $\alpha$.
        \item As $\xi\to1$, the factor $[L(\theta)]^{\frac{n}{n+n_0}}$ lessens the effect of contamination (since $n/(n+n_0)<1$).  This reduces $R_{\max}$ vs.\ $R_{\min}$, hence strictly decreases the final bound in $\xi$.
        \item Nonnegativity is obvious, and the bound remains finite and explicit for all parameter values.  
    \end{itemize}
    This completes the proof of the theorem.
\end{proof}

\begin{proof}[Proof for Corollary~\ref{cor:contamination_alpha}]
    By the definition of total variation distance,
    \begin{align*}
        d_{\mathrm{TV}}(p,q) =  \frac{1}{2} \int_{\Theta} \bigl|\,p(\theta)-q(\theta)\bigr|\,d\theta.
    \end{align*}
    We can rewrite each integrand $|p(\theta)-q(\theta)|$ by factoring out $q(\theta)$, since $q(\theta) \geq 0$,
    \begin{align*}
        \left|\,p(\theta)-q(\theta)\right| =  q(\theta)\,\left| \frac{p(\theta)}{q(\theta)} - 1 \right|.
    \end{align*}
    Hence
    \begin{align*}
        d_{\mathrm{TV}}(p,q) = \frac{1}{2}\int_{\Theta} q(\theta)\,\left|\frac{p(\theta)}{q(\theta)} - 1 \right| d\theta.
    \end{align*}
    We next observe that for any measurable function $g(\theta)$ taking real values,
    \begin{align*}
        \int_{\Theta} q(\theta)\,\bigl|\,g(\theta)\bigr|\;d\theta \leq  \Bigl(\sup_{\theta\in\Theta}\,\bigl|g(\theta)\bigr|\Bigr) \times \int_{\Theta} q(\theta)\,d\theta.
    \end{align*}
    But $\int_{\Theta} q(\theta)\,d\theta=1$ because $q$ is a probability density.  Therefore
    \begin{align*}
        \int_{\Theta} q(\theta)\,\bigl|\,g(\theta)\bigr|\;d\theta \leq  \sup_{\theta\in\Theta}\,\bigl|\,g(\theta)\bigr|.
    \end{align*}
    In our application, $g(\theta)=\frac{p(\theta)}{q(\theta)} -1$, so
    \begin{align*}
        \int_{\Theta} q(\theta)\,\left|\frac{p(\theta)}{q(\theta)}-1\right| \,d\theta \leq  \sup_{\theta\in\Theta}\,\left|\frac{p(\theta)}{q(\theta)}-1\right|.
    \end{align*}
    Consequently
    \begin{align*}
        d_{\mathrm{TV}}(p,q) = \frac{1}{2}\int q(\theta)\,\left|
    \frac{p(\theta)}{q(\theta)}-1 \right|\;d\theta \leq \frac{1}{2}\sup_{\theta\in\Theta}\,\left|\frac{p(\theta)}{q(\theta)}-1\right|.
    \end{align*}
    Recall the definitions in the statement of the theorem:
    \begin{align*}
        m = \inf_{\theta\in\Theta}\,\frac{p(\theta)}{q(\theta)}, \quad M = \sup_{\theta\in\Theta}\,\frac{p(\theta)}{q(\theta)}.
    \end{align*}
    Thus for every $\theta$,
    \begin{align*}
        m \;\le\; \frac{p(\theta)}{q(\theta)} \;\le\; M,
    \end{align*}
    and we want to bound
    \begin{align*}
        \sup_{\theta}\,\Bigl|\frac{p(\theta)}{q(\theta)}-1\Bigr|.
    \end{align*}
    Observe that for any real $x$,
    \begin{align*}
        \bigl|\,x-1\bigr| = 
       \begin{cases}
         x-1, &\text{if } x\ge1,\\
         1-x, &\text{if } x<1.
       \end{cases}
    \end{align*}
    Hence as $x$ ranges over $[\,m,\;M\,]$, the maximum deviation $|x-1|$ is
    \begin{align*}
        \max\!\bigl\{\;M-1,\;1-m\,\bigr\}.
    \end{align*}
    Therefore
    \begin{align*}
        \sup_{\theta}\,\left|\frac{p(\theta)}{q(\theta)} -1\right| &= \max\left\{\, \sup_\theta\,\left[\frac{p(\theta)}{q(\theta)} -1\right], \sup_\theta\,\left[1 - \frac{p(\theta)}{q(\theta)}\right]\right\} \\
        &= \max\!\bigl\{\,M -1,\;1 - m\bigr\}.
    \end{align*}
    Indeed,
    \begin{align*}
        \sup_\theta\;\Bigl(\frac{p(\theta)}{q(\theta)}-1\Bigr) = M-1,\quad \sup_\theta\;\Bigl(1-\frac{p(\theta)}{q(\theta)}\Bigr) = 1-m,
    \end{align*}
    so the overall supremum norm $\sup_\theta|\frac{p(\theta)}{q(\theta)}-1|$ is $\max\{\,M-1,\;1-m\}$.
    Putting them together,
    \begin{align*}
        d_{\mathrm{TV}}(p,q) \leq \frac12\,\sup_{\theta}\,\Bigl|\frac{p(\theta)}{q(\theta)}-1\Bigr| = \frac12\,\max\!\bigl\{\,M-1,\;1-m\bigr\}.
    \end{align*}
    That is precisely the statement of the theorem.
\end{proof}

\begin{proof}[Proof for Lemma~\ref{lem:derivative_analysis}]
    From the definition,
    \begin{align*}
        L_z(\theta) &= \frac{1}{z} \ln \left\{(1 - \xi) p_0(\theta)^z + \xi p_1(\theta)^z \right\} \\
        &= \frac{1}{z}\left\{p_0(\theta)^z\left((1 - \xi) + \xi\left(\frac{p_1(\theta)}{p_0(\theta)}\right)^z\right)\right\} \\
        &= \ln p_0(\theta) + \frac{1}{z}\ln\left\{(1 - \xi) + \xi R(\theta)^z \right\},
    \end{align*}
    where
    \begin{align*}
        R(\theta) = \frac{p_1(\theta)}{p_0(\theta)}.
    \end{align*}
    Hence,
    \begin{align*}
        L'_z(\theta) &= \frac{d}{d\theta}\ln p_0(\theta) + \frac{1}{z}\frac{d}{d\theta}\ln\left\{(1 - \xi) + \xi R(\theta)^z \right\} \\
        &= \frac{p'_0(\theta)}{p_0(\theta)} + \frac{1}{z}\frac{\xi z R(\theta)^{z-1}R'(\theta)}{(1 - \xi) + \xi R(\theta)^z} \\
        &= \frac{\xi R(\theta)^{z - 1} R'(\theta)}{(1 - \xi) + \xi R(\theta)^z}.
    \end{align*}
\end{proof}

\begin{proof}[Proof for Theorem~\ref{thm:shape_of_power_sum_posterior}]
    By assumption, $p_0$ and $p_1$ each have a single peak but located at different $m_0 < m_1$.
    Because $p_0$ and $p_1$ swap which is bigger on $(m_0, m_1)$, the ratio $R(\theta)$ is
    \begin{itemize}
        \item $R(\theta) < 1$ for $\theta \in (m_0, \theta^*)$,
        \item $R(\theta) = 1$ at $\theta = \theta^*$,
        \item $R(\theta) > 1$ for $\theta \in (\theta^*, m_1)$,
    \end{itemize}
    where $\theta^*$ is a some point.
    Also, $R(\theta) \ll 1$ near $m_0$, and $R(\theta) \gg 1$ near $m_1$.

    When $\alpha$ is large, $z = \frac{1 + \alpha}{2}$ is also large.
    Then, $R(\theta)^z$ has a threshold behavior.
    If $R(\theta) < 1$, then $R(\theta)^z \approx 0$, and if $R(\theta) > 1$ then $R(\theta)^z \approx \infty$.
    Hence, $\xi R(\theta)^z$ is nearly $0$ in the region where $p_1 < p_0$, and nearly infinite where $p_1 > p_0$.
    Consequently, for near $m_0$, we have $p_0(m_0) \gg p_1(m_0)$ and $R(m_0) \ll 1$, $\xi R(m_0)^z \approx 0$ as $z \to \infty$.
    Therefore, in a neighborhood of $m_0$, the bracket $(1 - \xi) + \xi R(\theta)^z \approx (1 - \xi) \neq 0$.
    So, let $F(\theta) \coloneqq \{(1 - \xi)p_0(\theta)^z + \xi p_1(\theta)^z \}^{1/z}$ and
    \begin{align*}
        F(\theta) \approx \left\{p(\theta^z)(1 - \xi)\right\}^{1/z}.
    \end{align*}
    The factor $(1 - \xi)^{1/z}$ is constant with respect to $\theta$.
    Hence, near $m_0$, $F(\theta)$ behaves essentially like $p_0(\theta)$ and we get a local maximum near $m_0$.
    Also, near $m_1$, we have $p_1(m_1) \gg p_0(m_1)$, so $R(m_1) \gg 1$, and $R(m_1)^z \to \infty$.
    Then, $\xi R(\theta)^z$ dominates and
    \begin{align*}
        F(\theta) &\approx \left(p_0(\theta)^z \xi R(\theta)^z\right)^{1/z} \\
        &= \left\{p_1(\theta)^z \left(\frac{p_0(\theta)}{p_1(\theta)}\right)^z \xi R(\theta)^z \right\}^{1/z} \\
        &= \left\{p_1(\theta)^z \xi \left(\frac{1}{R(\theta)}\right)^z\right\}^{1/z} \\
        &= \xi^{1/z}p_1(\theta)\left\{\frac{R(\theta)^z}{R(\theta)^z}\right\}^{1/z} \\
        &= \xi^{1/z} p_1(\theta).
    \end{align*}
    This again is just $p_1(\theta)$ up to a multiplicative constant.
    So there is a local maximum near $m_1$.

    Therefore, for large $\alpha$, there are two distinct neighborhoods, one near $m_0$ and one near $m_1$, in which $F(\theta)$ approximates $p_0$ and $p_1$ respectively.
    Since each $p_i$ has a strict mode, we conclude $F(\theta)$ has at least two local maxima.
    In particular, the normalized version $g(\theta)$ is multimodal.

    When $\alpha$ is small, specifically when $\alpha < 1$, then $z = \frac{1 + \alpha}{2} < 1$.
    Now the map $t \mapsto t^z$ is concave on $(0, \infty)$.
    As a result, the function $(p_0(\theta)^z + p_1(\theta)^z)$ is more merged than max-like.
    Concretely, by Lemma 4.3,
    \begin{align*}
        L'_z(\theta) = \frac{p'_0(\theta)}{p_0(\theta)} + \frac{\xi R(\theta)^{z-1}R'(\theta)}{(1 - \xi) + \xi R(\theta)^z},
    \end{align*}
    and because $z < 1$, the exponent $(z - 1)$ is negative.
    So if $p_1(\theta) > p_0(\theta)$, the ratio $(p_1(\theta) / p_0(\theta))^{z-1}$ is less dramatic. 
    This ensures only one sign change in $L'_z(\theta)$.
    This yields exavtly one local maximum overall.
    Hence, for smaller $\alpha$, we see unimodality, only one local maximum emerges.
\end{proof}

\begin{proof}[Proof for Theorem~\ref{thm:consistency}]
    Recall the definitions
    \begin{align*}
        p_0(\theta) = L(\theta \mid D)\pi_0(\theta), \quad p_1(\theta) = L(\theta \mid D)L(\theta \mid D_0)\pi_0(\theta).
    \end{align*}
    Also, let $z = (1 + \alpha) / 2$ and
    \begin{align*}
        R(\theta) &\coloneqq (1 - \xi)p_0(\theta)^z + \xi p_1(\theta)^z, \\
        \pi(\theta \mid D, D_0, \xi, \alpha) &= \frac{R(\theta)^{1/z}}{\int_\Theta R(\theta')^{1/z}d\theta'}.
    \end{align*}
    Let $C_0 \coloneqq R(\theta_0)$, which is strictly positive.
    Then,
    \begin{align*}
        \frac{R(\theta)}{R(\theta_0)} &= \frac{(1 - \xi)p_0(\theta)^z + \xi p_1(\theta)^z}{(1 - \xi)p_0(\theta_0) + \xi p_1(\theta_0)^z} = \frac{1}{C_0}\left\{(1 - \xi)p_0(\theta)^z + \xi p_1(\theta)^z\right\}, \\
        \ln \frac{R(\theta)}{R(\theta_0)} &= -\ln C_0 + \ln \left\{(1 - \xi)p_0(\theta)^z + \xi p_1(\theta)^z\right\} \eqqcolon \Delta_n(\theta).
    \end{align*}
    Recall that
    \begin{align*}
        p_0(\theta) &= L(\theta \mid D)\pi_0(\theta) = \left[\prod^n_{i=1}f(X_i ; \theta)\right]\times \pi_0(\theta), \\
        p_1(\theta) &= L(\theta \mid D)L(\theta \mid D_0)\pi_0(\theta) = \left[\prod^n_{i=1}f(X_i ; \theta)\right] \times \left[\prod^{n_0}_{j=1}f(Y_j; \theta)\right]\times \pi_0(\theta).
    \end{align*}
    Hence,
    \begin{align*}
        p_0(\theta)^z &= \exp\left\{z \ln (L(\theta \mid D)\pi_0(\theta))\right\} = \exp\left\{z\left[\ln L(\theta \mid D) + \ln \pi_0(\theta)\right]\right\}, \\
        p_1(\theta)^z &= \exp\left\{z \ln (L(\theta \mid D)L(\theta \mid D_0)\pi_0(\theta))\right\}.
    \end{align*}
    Let
    \begin{align*}
        A_n(\theta) &\coloneqq \ln\left(\frac{L(\theta \mid D)\pi_0(\theta)}{L(\theta_0 \mid D)\pi_0(\theta_0)}\right) = \ln\frac{\pi_0(\theta)}{\pi_0(\theta_0)} + \sum^n_{i=1}\ln\frac{f(X_i; \theta)}{f(X_i; \theta_0)}, \\
        B_{n_0}(\theta) &\coloneqq \ln\left(\frac{L(\theta \mid D_0)}{L(\theta_0 \mid D_0)}\right) = \sum^{n_0}_{j=1}\ln\frac{f(Y_j; \theta)}{f(Y_j; \theta_0)}.
    \end{align*}
    Then,
    \begin{align*}
        p_0(\theta)^z &= p_0(\theta_0)^z\exp\left\{z A_n(\theta)\right\}, \\
        p_1(\theta)^z &= p_1(\theta_0)^z\exp\left\{z A_n(\theta) + z B_{n_0}(\theta)\right\}, \\
        \Delta_n(\theta) &= -\ln C_0 + \ln \left\{a_0 e^{z A_n(\theta)} + b_0 e^{z\left[A_n(\theta) + B_{n_0}(\theta)\right]}\right\},
    \end{align*}
    where $a_0 = (1 - \xi)p_0(\theta)^z$ and $b_0 = \xi p_1(\theta)^z$, and hence $C_0 = a_0 + b_0$.
    For $A_n(\theta)$, by the Strong Law of Large Numbers (SLLN), we have
    \begin{align*}
        \frac{1}{n}\sum^n_{i=1}\ln\frac{f(X_i; \theta)}{f(X_i; \theta_0)} \overset{a.s.}{\underset{n\to\infty}{\to}} - D_{\mathrm{KL}}[\theta_0 \| \theta] < 0,
    \end{align*}
    whenever $\theta \neq \theta_0$.
    The additional term $\ln(\pi_0(\theta) / \pi_0(\theta_0))$ is constant in $n$ and hence negligible at order $n$.
    Therefore,
    \begin{align*}
        A_n(\theta) = O_{\theta_0}(n), \quad \frac{1}{n}A_n(\theta) \overset{a.s.}{\underset{n\to\infty}{\to}} -D_{\mathrm{KL}}[\theta_0 \| \theta].
    \end{align*}
    For $B_{n_0}(\theta)$, if $n_0 \to \infty$ similarly,
    \begin{align*}
        \frac{1}{n_0}\sum^{n_0}_{j=1}\ln\frac{f(Y_j ; \theta)}{f(Y_j; \theta_0)} \overset{a.s.}{\underset{n\to\infty}{\to}} -D_{\mathrm{KL}}[\theta_0 \| \theta] < 0.
    \end{align*}
    Thus, for every $\theta \neq \theta_0$, with probability going to $1$ as $n \to \infty$ and $n_0 \to \infty$,
    \begin{align*}
        z A_n(\theta) &\approx z n D_{\mathrm{KL}}[\theta_0 \| \theta] \to -\infty, \\
        z B_{n_0}(\theta) &\approx z n_0 D_{\mathrm{KL}}[\theta_0 \| \theta] \to -\infty.
    \end{align*}
    Hence the exponential terms $\exp\{z A_n(\theta)\}$ and $\exp\{z [A_n(\theta) + B_{n_0}(\theta)]\}$ both goes to $0$ as $n, n_0 \to \infty$, whenever $\theta \neq \theta_0$.

    Given that $a_0, b_0 > 0$ are constants and the exponential terms vanish exponentially fast for each fixed $\theta \neq \theta_0$, it follows that
    \begin{align*}
        (1 - \xi)p_o(\theta)^z + \xi p_1(\theta)^z = a_0 e^{z A_n(\theta)} + b_0 e^{z[A_n(\theta) + B_{n_0}(\theta)]} \overset{a.s.}{\underset{n\to\infty}{\to}} 0,
    \end{align*}
    for each fixed $\theta \neq \theta_0$.
    Fix $\epsilon > 0$.
    Consider the posterior mass of the set $\{\|\theta - \theta_0\| > \epsilon\}$,
    \begin{align*}
        \pi\left(\{\|\theta - \theta_0\| > \epsilon\} \mid D, D_0, \xi, \alpha\right) &= \frac{\int_{\|\theta - \theta_0\| > \epsilon}R(\theta)^{\frac{2}{1 + \alpha}}}{\int_\Theta R(\theta)'^{\frac{2}{1+\alpha}}d\theta'} \\
        &= \frac{R(\theta_0)^{\frac{2}{1+\alpha}}\int_{\|\theta - \theta_0\| > \epsilon}\left(\frac{R(\theta)}{R(\theta_0)}\right)^{\frac{2}{1+\alpha}}}{R(\theta_0)^{\frac{2}{1+\alpha}}\int_\Theta\left(\frac{R(\theta')}{R(\theta_0)}\right)^{\frac{2}{1+\alpha}}} \\
        &= \frac{\int_{\|\theta - \theta_0\| > \epsilon}\left(\frac{R(\theta)}{R(\theta_0)}\right)^{\frac{2}{1+\alpha}}}{\int_\Theta\left(\frac{R(\theta')}{R(\theta_0)}\right)^{\frac{2}{1+\alpha}}} \\
        &\leq \sup_{\|\theta - \theta_0\| > \epsilon}\left(\frac{R(\theta)}{R(\theta_0)}\right)^{\frac{2}{1+\alpha}} \overset{P_{\theta_0}}{\underset{n,n_0\to\infty}{\to}} 0.
    \end{align*}
    This concludes the proof.
\end{proof}

\begin{proof}[Proof for Theorem~\ref{thm:asymptotic_variance}]
    Let $z = (1 + \alpha) / 2$, and
    \begin{align*}
        R(\theta) &= (1 - \xi) p_0(\theta)^z + \xi p_1(\theta)^z, \\
        \pi(\theta \mid D, D_0, \alpha, \xi) &\propto R(\theta)^{1/z} = \exp\left\{Q(\theta)\right\},
    \end{align*}
    where $Q(\theta) = \frac{1}{z}\ln R(\theta)$.
    Also, let $\hat{\theta}_n = \argmax_\theta Q(\theta)$ be the posterior mode, and $h \coloneqq \sqrt{n}(\theta - \theta_0)$,
    \begin{align*}
        A_n(\theta) &\coloneqq \ln\frac{p_0(\theta)}{p_0(\theta_0)} = \sum^n_{i=1}\frac{f(X_i; \theta)}{f(X_i; \theta_0)} + \ln\frac{\pi_0(\theta)}{\pi_0(\theta_0)}, \\
        B_{n_0}(\theta) &\coloneqq \ln\frac{L(\theta \mid D_0)}{L(\theta_0 \mid D_0)} = \sum^{n_0}_{j=1}\ln\frac{f(Y_j; \theta)}{f(Y_j; \theta_0)}.
    \end{align*}
    Then,
    \begin{align*}
        p_0(\theta)^z &= p_0(\theta)^z\exp\left\{z A_n(\theta)\right\}, \\
        p_1(\theta)^z &= p_1(\theta_0)^z\exp\left\{z\left[A_n(\theta) + B_{n_0}(\theta)\right]\right\}, \\
        R(\theta) &= (1 - \xi) p_0(\theta_0)^z e^{z A_n(\theta)} + \xi p_1(\theta_0)^z e^{z[A_n(\theta) + B_{n_0}(\theta)]}.
    \end{align*}
    Write $A_n(\theta_0 + h / \sqrt{n})$ as
    \begin{align*}
        A_n\left(\theta_0 + \frac{h}{\sqrt{n}}\right) &= A_n(\theta_0) + \frac{h^\top}{\sqrt{n}}\nabla_\theta A_n(\theta_0) + \frac{1}{2n}h^\top\nabla_\theta^2A_n(\theta_0)h \\
        &\quad\quad\quad + \frac{1}{6n^{3/2}}\sum_{r,s,t,}M_{rst}h_rh_sh_t + \cdots,
    \end{align*}
    where $M_{rst} = \frac{\partial^3}{\partial\theta_r\partial\theta_s\partial\theta_t}A_n(\theta_0)$.
    Simlarly,
    \begin{align*}
        B_{n_0}\left(\theta_0 + \frac{h}{\sqrt{n}}\right) &= B_{n_0}(\theta_0) + \frac{h^\top}{\sqrt{n}}\nabla_\theta B_{n_0}(\theta_0) + \frac{1}{2n}h^\top \nabla_\theta^2 B_{n_0}(\theta_0)h \\
        &\quad\quad\quad + \frac{1}{6n^{3/2}}\sum_{r,s,t}N_{rst}h_rh_sh_t + \cdots,
    \end{align*}
    with $N_{rst} = \frac{\partial^3}{\partial\theta_r\partial\theta_s\partial\theta_t} B_{n_0}(\theta_0)$.
    Under the true model,
    \begin{align*}
        A_n(\theta_0) = 0, \quad B_{n_0}(\theta_0) = 0.
    \end{align*}
    Hence, expansions become
    \begin{align*}
        A_n\left(\theta_0 + \frac{h}{\sqrt{n}}\right) &= \frac{h^\top}{\sqrt{n}}\nabla_\theta A_n(\theta_0) + \frac{1}{2n}h^\top \nabla_\theta^2 A_n(\theta)h + \frac{1}{6n^{3/2}}\sum_{r,s,t,}M_{rst}h_rh_sh_t + \cdots, \\
        B_{n_0}\left(\theta_0 + \frac{h}{\sqrt{n}}\right) &= \frac{h^\top}{\sqrt{n}}\nabla_\theta B_{n_0}(\theta_0) + \frac{1}{2n}h^\top \nabla_\theta^2 B_{n_0}(\theta_0)h + \frac{1}{6n^{3/2}}\sum_{r,s,t}N_{rst}h_rh_sh_t + \cdots.
    \end{align*}
    We can write as
    \begin{align*}
        Q\left(\theta_0 + \frac{h}{\sqrt{n}}\right) &= \frac{1}{z}\ln\left\{a_0 e^u + b_0 e^{u + w}\right\} \\
        &= \frac{1}{z}\ln\left\{e^u(a_0 + b_0 e^w)\right\} \\
        &= \frac{1}{z}\left[u + \ln(a_0 + b_0e^w)\right],
    \end{align*}
    where
    \begin{align*}
        u &= \frac{z}{\sqrt{n}}h^\top g_n + \frac{z}{2n}h^\top H_n h + \cdots, \\
        w &= \frac{z}{\sqrt{n}}h^\top g_{n_0} + \frac{z}{2n}h^\top H_{n_0}h + \cdots, \\
        g_n &= \nabla_\theta A_n(\theta_0), \quad g_{n_0} = \nabla_\theta B_{n_0}(\theta_0), \\
        H_n &= \nabla_\theta^2 A_n(\theta_0), \quad H_{n_0} = \nabla^2_\theta B_{n_0}(\theta_0), \\
        a_0 &= (1 - \xi)p_0(\theta)^z, \quad b_0 = \xi p_1(\theta)^z.
    \end{align*}
    We next expand $\ln(a_0 + b_0 e^w)$ and we have
    \begin{align*}
        Q\left(\theta_0 + \frac{h}{\sqrt{n}}\right) &= \frac{1}{z}\left[u + \ln(a_0 + b_0) + \frac{b_0}{a_0 + b_0}w + \frac{a_0b_0}{2(a_0 + b_0)^2}w^2 + \cdots\right] \\
        &= \frac{1}{z}u + \frac{1}{z}\ln(a_0 + b_0) + \frac{b_0}{z(a_0 + b_0)} + \frac{a_0b_0}{2z(a_0 + b_0)}w^2 + \cdots.
    \end{align*}
    Let $\hat{h}_n = \sqrt{n}(\hat{\theta}_n - \theta_0)$ and one obtains the expansion as
    \begin{align*}
        \hat{h} = H^{-1}\left[g_n + \frac{b_0}{a_0 + b_0}g_{n_0}\right] + H^{-1}\left[\cdots\right]H^{-1} + O_p(n^{-1/2}),
    \end{align*}
    where $H$ is the Hessian term that includes partial dependence on $\alpha$:
    \begin{align*}
        H = -\nabla_\theta^2Q(\theta) |_{\theta = \theta_0} \approx n I(\theta_0) + \xi n_0 I_0(\theta_0).
    \end{align*}
    Therefore,
    \begin{align*}
        \mathrm{Var}(\hat{\theta}_n) = \frac{1}{n}\left[I(\theta_0) + \frac{\xi n_0}{n}I_0(\theta_0)\right]^{-1} + \frac{1}{n^2}\Gamma_{\alpha,\xi} + o\left(\frac{1}{n^2}\right),
    \end{align*}
    where
    \begin{align*}
        \Gamma_{\alpha,\xi} = \Sigma_0\left\{\frac{1}{2}\nabla_\theta^3Q(\theta_0)[\Sigma_0, \Sigma_0, \Sigma_0] - S(\theta_0)\right\},
    \end{align*}
    where $S(\theta_0)$ is the collection of mode‐shift terms, and
    \begin{align*}
        \Sigma_0 = \left[-\nabla_\theta^2 Q(\theta_0)\right]^{-1} \approx \left[n I(\theta_0) + \xi n_0 I_0(\theta_0)\right]^{-1}.
    \end{align*}
    Here,
    \begin{align*}
        \nabla^3_\theta Q(\theta_0) &= \nabla_\theta^3 A_n(\theta_0) + \left.\frac{1}{z}\nabla_\theta^3\left[\ln \left(a_0 + b_0 e^{z B_{n_0}(\theta)}\right)\right] \right|_{\theta = \theta_0} \\
        &= \nabla_\theta^3 A_n(\theta_0) + \frac{b_0\left[z\nabla^3_\theta B_{n_0}(\theta_0) + 3 z^2(H_{n_0}\otimes g_{n_0}) + z^3(g_{n_0})^{\otimes 3}\right]}{z(a_0 + b_0)} \\
        &\quad\quad\quad - 3\frac{b_0\left[zH_{n_0} + z^2(g_{n_0}\otimes g_{n_0})\right] \ \otimes [b_0 z g_{n_0}]}{z(a_0 + b_0)^2} + \frac{2[b_0 z g_{n_0}]^{\otimes 3}}{z(a_0 + b_0)^3}.
    \end{align*}
\end{proof}

\section{Derivation of Explicit Examples}
\label{sec:derivation_of_explicit_examples}

\begin{example}[Univeriate Gaussian]
    Let $D = \{X_1,\dots,X_n\}$, $X_i \mid \theta \sim \mathcal{N}(\theta, \sigma^2)$ and $D_0 = \{Y_1,\dots,Y_n\}$, $Y_j \mid \theta \sim \mathcal{N}(\theta, \sigma^2)$ be the current and historical data.
    Also $\pi_0(\theta) = \mathcal{N}(\mu_0, \tau_0^2)$ be the baseline prior.
    Then, for $\alpha \neq \pm 1$ and $\xi \in [0, 1]$, the generalized power posterior $\theta \mapsto g(\theta)$ is given as
    \begin{align*}
       g(\theta) \propto \exp\left\{-\frac{1}{2}\left(\frac{n}{\sigma^2} + \frac{1}{\tau_0^2}\right)\theta + \left(\frac{S_X}{\sigma^2} + \frac{\mu_0}{\tau_0^2}\right)\theta\right\}\left\{1 + C\exp\left(\frac{z S_Y}{\sigma^2}\theta - \frac{z n_0}{2\sigma^2}\theta^2\right)\right\}^{1/z}.
   \end{align*}
    where $S_X = \sum^n_{i=1}X_i$, $S_Y = \sum^{n_0}_{j=1}Y_j$, and
    \begin{align*}
        C &= \frac{Q_1}{Q_0}\exp\left[-\frac{z}{2\sigma^2}\sum^{n_0}_{j=1}Y_j^2\right], \\
        Q_0 &= (1 - \xi)\frac{1}{(2\pi\sigma^2)^{\frac{nz}{2}}}\frac{1}{(2\pi\tau_0^2)^{\frac{z}{2}}}, \\
        Q_1 &= \xi\frac{1}{(2\pi\sigma^2)^{\frac{(n + n_0)z}{2}}}\frac{1}{(2\pi\tau_0^2)^{\frac{z}{2}}}.
    \end{align*}
    The normalization constant is determined by integrating with respect to $\theta$.
\end{example}
\begin{proof}
   First, we have
   \begin{align*}
       L(\theta \mid D) &= \prod^n_{i=1}\frac{1}{\sqrt{2\pi\sigma^2}}\exp\left(-\frac{(X_i - \theta)^2}{2\sigma^2}\right), \\
       L(\theta \mid D_0) &= \prod^{n_0}_{j=1}\frac{1}{\sqrt{2\pi\sigma^2}}\exp\left(-\frac{(Y_j - \theta)^2}{2\sigma^2}\right), \\
       \pi_0(\theta) &= \frac{1}{\sqrt{2\pi\tau_0^2}}\exp\left(-\frac{(\theta - \mu_0)}{2\tau_0^2}\right).
   \end{align*}
   Then,
   \begin{align*}
       p_0(\theta) &= L(\theta \mid D)\pi_0(\theta) \\
       &= \left\{\prod^n_{i=1}\frac{1}{\sqrt{2\pi\sigma^2}}\exp\left(-\frac{(X_i - \theta)}{2\sigma^2}\right)\right\} \times \left\{\frac{1}{\sqrt{2\pi\sigma^2}}\exp\left(-\frac{(\theta - \mu_0)^2}{2\tau_0^2}\right)\right\} \\
       &= \frac{1}{(2\pi\sigma^2)^{\frac{n}{2}}}\exp\left\{-\frac{1}{2\sigma^2}\sum^n_{i=1}(X_i - \theta)^2 \right\} \times \frac{1}{\sqrt{2\pi\tau_0^2}}\exp\left\{-\frac{(\theta - \mu_0)^2}{2\tau_0^2}\right\}, \\
       p_0(\theta)^z &= \frac{1}{(2\pi\sigma^2)^{\frac{nz}{2}}}\exp\left\{-\frac{z}{2\sigma^2}\sum^n_{i=1}(X_i - \theta)^2 \right\} \times \frac{1}{(2\pi\tau_0^2)^{\frac{z}{2}}}\exp\left\{-z\frac{(\theta - \mu_0)^2}{2\tau_0^2}\right\}, \\
       p_1(\theta) &= L(\theta \mid D)L(\theta \mid D_0)\pi_0(\theta) \\
       &= \left\{\prod^n_{i=1}\frac{1}{\sqrt{2\pi\sigma^2}}\exp\left(-\frac{(X_i - \theta)^2}{2\sigma^2}\right)\right\} \times \left\{\prod^{n_0}_{j=1}\frac{1}{\sqrt{2\pi\sigma^2}}\exp\left(-\frac{(Y_j - \theta)^2}{2\sigma^2}\right)\right\} \\
       &\quad\quad\quad\quad\quad\quad \times \frac{1}{\sqrt{2\pi\sigma^2}}\exp\left(-\frac{(\theta - Y_j)^2}{2\tau_0^2}\right) \\
       &= \frac{1}{(2\pi\sigma^2)^{\frac{n + n_0}{2}}}\exp\left\{-\frac{1}{2\sigma^2}\sum^n_{i=1}(X_i - \theta)^2 - \frac{1}{2\sigma^2}\sum^{n_0}_{j=1}(Y_j - \theta)^2\right\} \\
       &\quad\quad\quad\quad\quad\quad \times \frac{1}{\sqrt{2\pi\sigma^2}}\exp\left(-\frac{(\theta - Y_j)^2}{2\tau_0^2}\right), \\
       p_1(\theta)^z &= \frac{1}{(2\pi\sigma^2)^{\frac{(n + n_0)z}{2}}}\exp\left\{-\frac{z}{2\sigma^2}\sum^n_{i=1}(X_i - \theta)^2 - \frac{z}{2\sigma^2}\sum^{n_0}_{j=1}(Y_j - \theta)^2\right\} \\
       &\quad\quad\quad\quad\quad\quad \times \frac{1}{(2\pi\tau_0^2)^{\frac{z}{2}}}\exp\left(-z\frac{(\theta - Y_j)^2}{2\tau_0^2}\right).
   \end{align*}
   Hence, we can write
   \begin{align*}
       p_0(\theta)^z &= \tilde{Q}_0 \exp\left(-z A_0(\theta) \right), \\
       p_1(\theta)^z &= \tilde{Q}_1 \exp\left(-z A_1(\theta) \right),
   \end{align*}
   where
   \begin{align*}
       \tilde{Q}_0 &= \frac{1}{(2\pi\sigma^2)^{\frac{nz}{2}}(2\pi\tau_0^2)^{\frac{z}{2}}}, \\
       \tilde{Q}_1 &= \frac{1}{(2\pi\sigma^2)^{\frac{(n + n_0)z}{2}}(2\pi\tau_0^2)^{\frac{z}{2}}}, \\
       A_0(\theta) &= \frac{1}{2\sigma^2}\sum^n_{i=1}(X_i - \theta)^2 + \frac{(\theta - \mu_0)^2}{2\sigma^2}, \\
       A_1(\theta) &= \frac{1}{2\sigma^2}\sum^n_{i=1}(X_i - \theta)^2 + \frac{1}{2\sigma^2}\sum^{n_0}_{j=1}(Y_j - \theta) + \frac{(\theta - \mu_0)^2}{2\tau_0^2}.
   \end{align*}
   Then,
   \begin{align*}
       (1 - \xi) p_0(\theta)^z + \xi p_1(\theta)^z &= (1 - \xi)\tilde{Q}_0 \exp(-z A_0(\theta)) + \xi \tilde{Q}_1 \exp(-z A_1(\theta)) \\
       &= \exp(-z A_0(\theta))\left\{(1 - \xi)\tilde{Q}_0 + \xi \tilde{Q}_1 \exp(-z(A_1(\theta) - A_0(\theta)))\right\}, \\
       g(\theta) &\propto \exp(-A_0 \theta)\left\{(1 - \xi)\tilde{Q}_0 + \xi \tilde{Q}_1 \exp(-z(A_1(\theta) - A_0(\theta)))\right\}^{\frac{1}{z}}.
   \end{align*}
   Since
   \begin{align*}
       A_1(\theta) - A_0(\theta) = \frac{1}{2\sigma^2}\sum^{n_0}_{j=1}(Y_j - \theta)^2,
   \end{align*}
   we have
   \begin{align*}
       \exp\left\{-z(A_1(\theta) - A_0(\theta))\right\} = \exp\left\{-z\cdot \frac{1}{2\sigma^2}\sum^{n_0}_{j=1}(Y_j - \theta)^2\right\}.
   \end{align*}
   Define
   \begin{align*}
       S_X &= \sum^n_{i=1}X_i, \\
       S_Y &= \sum^{n_0}_{j=1}Y_j.
   \end{align*}
   Then, we expand each squared term as
   \begin{align*}
       \sum^n_{i=1}(X_i - \theta)^2 &= \sum^n_{i=1}X_i^2 - 2\theta S_X + n\theta^2, \\
       (\theta - \mu_0)^2 &= \theta^2 - 2\mu_0\theta + \mu_0^2, \\
       \sum^{n_0}_{j=1}(Y_j - \theta)^2 &= \sum^{n_0}_{j=1}Y_j^2 - 2\theta S_Y + n_0 \theta^2.
   \end{align*}
   Since factors that do not involve $\theta$ can be absorbed into the proportionality constant, we drop all terms independed of $\theta$.
   Thus, the parts depending on $\theta$ become
   \begin{align*}
       \exp\left\{\frac{\theta S_X}{\sigma^2} - \frac{n\theta^2}{2\sigma^2} + \frac{\mu_0\theta}{\tau_0^2} - \frac{\theta^2}{2\tau_0^2}\right\} = \exp\left\{-\frac{1}{2}\left(\frac{1}{\sigma^2} + \frac{1}{\tau_0^2}\right)\theta^2 + \left(\frac{S_X}{\sigma^2} + \frac{\mu_0}{\tau_0^2}\right)\theta \right\},
   \end{align*}
   and
   \begin{align*}
       &Q_0 + Q_1\exp\left[-\frac{z}{2\sigma^2}\left(\sum^{n_0}_{j=1}Y_j^2 - 2\theta S_Y + n_0\theta^2 \right)\right] \\
       &\quad\quad\quad = Q_0 + Q_1\exp\left[-\frac{z}{2\sigma^2}\sum^{n_0}_{j=1}Y_j\right]\exp\left\{\frac{z\theta S_Y}{\sigma^2} - \frac{zn_0\theta^2}{2\sigma^2}\right\} \\
       &\quad\quad\quad \propto \left\{1 + C\exp\left(\frac{z\theta S_Y}{\sigma^2} - \frac{z n_0\theta^2}{2\sigma^2}\right)\right\}^{1/z},
   \end{align*}
   where
   \begin{align*}
       C = \frac{Q_1}{Q_0}\exp\left[-\frac{z}{2\sigma^2}\sum^{n_0}_{j=1}Y_j^2\right].
   \end{align*}
   Putting these together, we obtain
   \begin{align*}
       g(\theta) \propto \exp\left\{-\frac{1}{2}\left(\frac{n}{\sigma^2} + \frac{1}{\tau_0^2}\right)\theta + \left(\frac{S_X}{\sigma^2} + \frac{\mu_0}{\tau_0^2}\right)\theta\right\}\left\{1 + C\exp\left(\frac{z S_Y}{\sigma^2}\theta - \frac{z n_0}{2\sigma^2}\theta^2\right)\right\}^{1/z}.
   \end{align*}
\end{proof}

\begin{example}[Beta–Bernoulli]
    Let $D = \{X_1,\dots,X_n\}$ with $X_i \in \{0, 1\}$ and $D_0 = \{Y_1,\dots,Y_{n_0}\}$ with $Y_j \in \{0, 1\}$ be the current and historical data from Bernoulli distributions with parameter $\theta \in (0, 1)$.
    Also, suppose that the baseline prior is $\pi_0(\theta) = \mathrm{Beta}(\alpha_0, \beta_0)$ for $\alpha_0, \beta_0 > 0$.
    Then, the corresponding generalized power posterior is given as
    \begin{align*}
        g(\theta) \propto \left\{F(\theta)\right\}^{\frac{1}{z}},
    \end{align*}
    where $z = \frac{1 + \alpha}{2}$
    \begin{align*}
        F(\theta) &= (1 - \xi)\left\{\theta^{\sum^n_{i=1}X_i + \alpha_0 - 1}(1 - \theta)^{n - \sum^n_{i=1}X_i + \beta_0 - 1}\right\}^z \\
        &\quad\quad\quad\quad\quad + \xi\left\{\theta^{\sum^n_{i=1}X_i + \sum^{n_0}_{j=1}Y_j + \alpha_0 - 1}(1 - \theta)^{n + n_0 - \sum^n_{i=1}X_i - \sum^{n_0}_{j=1}Y_j + \beta_0 - 1}\right\}^z
    \end{align*}
    This is generally not Beta anymore unless $\alpha = 1$ or $\xi=0,1$.
\end{example}
\begin{proof}
   Since
   \begin{align*}
       L(\theta \mid D) &= \theta^{\sum^n_{i=1}X_i}(1 - \theta)^{n - \sum^n_{i=1}X_i}, \\
       L(\theta \mid D_0) &= \theta^{\sum^{n_0}_{j=1}Y_j}(1 - \theta)^{n_0 \sum^{n_0}_{j=1}Y_j}, \\
       \pi_0(\theta) &= \mathrm{Beta}(\alpha_0, \beta_0),
   \end{align*}
   simple calculation yields
   \begin{align*}
       p_0(\theta) &= \theta^{\sum^{n}_{i=1}X_i}(1 - \theta)^{n - \sum^n_{i=1}X_i}\theta^{\alpha_0 - 1}(1-\theta)^{\beta_0 - 1} \\
       &= \theta^{\sum^n_{i=1}X_i + \alpha_0 - 1}(1 - \theta)^{n - \sum^n_{i=1}X_i + \beta_0 - 1}, \\
       p_1(\theta) &= \theta^{\sum^n_{i=1}X_i}(1 - \theta)^{n - \sum^n_{i=1}X_i}\theta^{\sum^{n_0}_{j=1}Y_j}(1 - \theta)^{n_0 - \sum^{n_0}_{j=1}Y_j}\theta^{\alpha_0 - 1}(1 - \theta)^{\beta_0 - 1} \\
       &= \theta^{\sum^n_{i=1}X_i + \sum^{n_0}_{j=1}Y_j + \alpha_0 - 1}(1 - \theta)^{n + n_0 - \sum^n_{i=1}X_i - \sum^{n_0}_{j=1}Y_j + \beta_0 - 1}.
   \end{align*}
   Let
   \begin{align*}
       S_X = \sum^n_{i=1}X_i, \quad S_Y = \sum^{n_0}_{j=1}Y_j.
   \end{align*}
   Then,
   \begin{align*}
       F(\theta) &= (1 - \xi)\left\{\theta^{S_X + \alpha_0 - 1}(1 - \theta)^{n - S_X + \beta_0 - 1}\right\}^z \\
       &\quad\quad\quad\quad + \xi\left\{\theta^{S_X + S_Y + \alpha_0 - 1}(1 - \theta)^{n + n_0 - S_X - S_Y + \beta_0 - 1}\right\}^z
   \end{align*}
   Notice that the $\theta$-dependent part in the first term is
   \begin{align*}
        \theta^{S_X + \alpha_0 - 1}(1 - \theta)^{n - S_X + \beta_0 - 1}.
   \end{align*}
   Since this is common in both terms, we factor it out as
   \begin{align*}
       F(\theta) &= \theta^{z(S_X + \alpha_0 - 1)}(1 - \theta)^{z(n - S_X + \beta_0 - 1)} \times \left\{(1 - \xi) + \xi \theta^{z S_Y}(1 - \theta)^{z(n_0 - S_Y)}\right\}.
   \end{align*}
   Then,
   \begin{align*}
       g(\theta) &\propto \left\{F(\theta)\right\}^{1/z} \\
       &= \theta^{S_X + \alpha_0 - 1}(1 - \theta)^{n - S_X + \beta_0 - 1} \times \left\{(1 - \xi) + \xi \theta^{z S_Y}(1 - \theta)^{z(n_0 - S_Y)}\right\}^{1/z}.
   \end{align*}
\end{proof}

\begin{example}[Dirichlet–Multinomial]
    Let $D = \{X_1,\dots,X_n\}$ with $\sum^n_{i=1 }X_i = n$ and $D_0 = \{Y_1,\dots,Y_{n_0}\}$ with $\sum^{n_0}_{j=1}Y_j = n_0$ from the multinomial distribution having the parameter $\bm{\theta} = (\theta_1,\theta_2,\dots,\theta_k)$.
    Also, suppose that the baseline prior is $\pi_0(\bm{\theta}) = \mathrm{Dirichlet}(\alpha_{0,1},\alpha_{0,2},\dots,\alpha_{0,k})$.
    Then, the corresponding generalized power posterior is given as
    \begin{align*}
        g(\bm{\theta}) \propto \left\{(1 - \xi)\prod^k_{i=1}\theta_i^{z(X_i + \alpha_{0,i} - 1)} + \xi \prod^k_{j=1}\theta_i^{z(X_i + Y_j + \alpha_{0,j} - 1)} \right\}^{\frac{1}{z}},
    \end{align*}
    where $z = \frac{1 + \alpha}{2}$.
    This does not yield a simple Dirichlet form, unless $\alpha = 1$, that is a non-Dirichlet shape on the simplex.
\end{example}
\begin{proof}
    The multinomial likelihood for $\bm{\theta} = (\theta_1,\theta_2,\dots,\theta_k)$ and the Dirichlet prior are given as
    \begin{align*}
        L(\bm{\theta} \mid D) &= n!\prod^k_{i=1}\theta_i^{X_i}, \\
        L(\bm{\theta} \mid D_0) &= n_0!\prod^k_{j=1}\theta_j^{Y_j}, \\
        \pi_0(\bm{\theta}) &\propto \prod^k_{i=1}\theta^{\alpha_{0,i} - 1}.
    \end{align*}
    Then,
    \begin{align*}
        p_0(\bm{\theta}) &\propto \left(\prod^k_{i=1}\theta_i^{X_i}\right)\left(\prod^k_{i=1}\theta^{\alpha_{0,i} - 1}\right) = \prod^k_{i=1}\theta_i^{X_i + \alpha_{0,i} - 1}, \\
        p_1(\bm{\theta}) &\propto \left(\prod^k_{i=1}\theta_i^{X_i}\right)\left(\prod^k_{j=1}\theta_j^{Y_j}\right)\left(\prod^k_{i=1}\theta^{\alpha_{0,i} - 1}\right) = \prod^k_{i=1}\theta_i^{X_i + Y_i + \alpha_{0,i} - 1}.
    \end{align*}
    Therefore,
    \begin{align*}
        g(\bm{\theta}) &\propto \left\{ (1 - \xi) p_0(\bm{\theta}) + \xi p_1(\bm{\theta}) \right\}^{\frac{1}{z}} \\ 
        &= \left\{(1 - \xi) \left(\prod^k_{i=1}\theta_i^{X_i + \alpha_{0,i} - 1}\right)^z + \xi\left(\prod^k_{i=1}\theta_i^{X_i + Y_i + \alpha_{0,i} - 1}\right)^z \right\}^{\frac{1}{z}} \\
        &= \left\{(1 - \xi)\prod^k_{i=1}\theta_i^{z(X_i + \alpha_{0,i} - 1)} + \xi \prod^k_{j=1}\theta_i^{z(X_i + Y_j + \alpha_{0,j} - 1)} \right\}^{\frac{1}{z}}.
    \end{align*}
\end{proof}

\end{document}